\documentclass[a4paper, 12pt, titlepage=false]{article}
\usepackage[english]{babel}
\usepackage[utf8]{inputenc}
\usepackage[T1]{fontenc}
\usepackage{setspace}
\usepackage{datetime}
\usepackage{mathptmx}
\usepackage{graphicx}
\usepackage[utf8]{inputenc}
\usepackage[top=2.5cm, left=2.5cm, bottom=2.5cm, right=2.5cm]{geometry} 
\usepackage{amsmath,amsthm,amssymb}
\usepackage{url} 
\usepackage{csquotes}
\usepackage{natbib}
\usepackage{indentfirst}
\usepackage[colorlinks = true, linkcolor = black, citecolor=black, urlcolor=black, filecolor = black, bookmarks=false,hypertexnames=false]{hyperref}
\usepackage[none]{hyphenat} 
\bibpunct{\textcolor{black}{(}}{\textcolor{black}{)}}{;}{a}{}{,}
\usepackage{float}
\usepackage{comment}
\usepackage{colortbl}
\usepackage{subcaption} 
\usepackage{caption} 
\usepackage{titling} 
\usepackage{hyperref}
\usepackage{booktabs}
\usepackage{algorithm}
\usepackage{algpseudocode}
\usepackage{url}
\usepackage{threeparttable}

\usepackage{times}
\newcommand{\YourTitleReport}{Understanding AI Evaluation Patterns:}
\newcommand{\YourSubtitle}{How Different GPT Models Assess Vision-Language Descriptions}

\begin{document}

\begin{center}
\large\textbf{\YourTitleReport} \\
\textbf{\YourSubtitle}

\vspace{0.5cm}
\normalsize
\textbf{Sajjad Abdoli, Rudi Cilibrasi, Rima Al-Shikh\textsuperscript{1}}

\vspace{0.2cm}
\small
\textsuperscript{1}Perle.ai, \{sajjad, rima, rudi.cilibrasi\}@perle.ai\\
\end{center}

\pagenumbering{arabic}
\setcounter{page}{1}

\begin{center}
\section*{Abstract}
\end{center}
\begingroup
\small 
\setlength{\parindent}{0pt} 
\sloppy
As AI systems increasingly evaluate other AI outputs, understanding their assessment behavior becomes crucial for preventing cascading biases. This study analyzes vision-language descriptions generated by NVIDIA's Describe Anything Model and evaluated by three GPT variants (GPT-4o, GPT-4o-mini, GPT-5) to uncover distinct "evaluation personalities" the underlying assessment strategies and biases each model demonstrates. GPT-4o-mini exhibits systematic consistency with minimal variance, GPT-4o excels at error detection, while GPT-5 shows extreme conservatism with high variability. Controlled experiments using Gemini 2.5 Pro as an independent question generator validate that these personalities are inherent model properties rather than artifacts. Cross-family analysis through semantic similarity of generated questions reveals significant divergence: GPT models cluster together with high similarity while Gemini exhibits markedly different evaluation strategies. All GPT models demonstrate a consistent 2:1 bias favoring negative assessment over positive confirmation, though this pattern appears family-specific rather than universal across AI architectures. These findings suggest that evaluation competence does not scale with general capability and that robust AI assessment requires diverse architectural perspectives.
\endgroup

\section{Introduction}

AI systems now routinely evaluate the outputs of other AI systems from code reviews to content moderation to training data curation. This recursive evaluation creates a critical dependency: biases and limitations in AI evaluators can amplify through successive generations of models trained on AI-assessed data. Yet the fundamental understanding of how different AI architectures conceptualize quality, what systematic biases they exhibit, and whether architectural advances improve or degrade evaluation consistency has not been studied extensively.

This study tackles the challenge by systematically analyzing AI evaluation behavior through 762 image descriptions generated by NVIDIA’s state-of-the-art DAM model. These outputs were assessed by three GPT-based judges GPT-4o, GPT-4o-mini, and GPT-5. Our analysis uncovers distinct “evaluation personalities” in each model, revealing their characteristic assessment strategies, inherent biases, and areas of emphasis when performing evaluation tasks.

NVIDIA's Describe Anything Model, provides the perfect foundation for evaluation analysis. The 3-billion parameter model achieves 67.3\% accuracy on DLC-Bench, a new technique for assessing the quality of AI generated descriptions, and employs innovative focal prompt technology that combines full image context with high-resolution crops of target regions~\citep{Lian2025DescribeAnything}. DAM addresses the fundamental challenge of generating detailed and accurate descriptions for specific regions in images and videos through two key innovations: the \textbf{focal prompt}, which ensures high-resolution encoding of targeted regions while preserving global context, and the \textbf{localized vision backbone}, which integrates precise localization with broader context using gated cross-attention adapters.

The landscape of AI-based evaluation has undergone fundamental transformation recently. Research shows that LLM-based assessment models like CLAIR achieve correlations of 0.604 to 0.627 with human judgment on image caption evaluation, outperforming traditional metrics like SPICE (0.449) and approaching but not reaching inter-human agreement levels (0.736). While these models show strong performance, they face challenges with non-determinism and occasional parsing failures, though variance between runs remains relatively small.~\citep{Chen2024MLLMJudge}. This variability suggests that different AI models may emphasize different aspects of assessment, making it crucial to understand their evaluation patterns rather than simply measuring their accuracy.

Large language models (LLMs) and multimodal large language models (MLLMs) have shown considerable promise as evaluators, or "LLM-as-a-Judge," providing scalable, flexible assessment capabilities across modalities~\citep{Chen2024MLLMJudge, Wei2022ChainOfThought}. However, studies reveal that despite impressive human-like discernment in pairwise comparisons, significant challenges persist, including biases, hallucinations, and inconsistencies, particularly in scoring and batch ranking tasks~\citep{Chen2024MLLMJudge}. These findings highlight the need for rigorous evaluation frameworks and bias-aware assessment methodologies tailored to LLM and MLLM judging capabilities.

Our methodology employs multiple experimental controls to ensure robust findings. In addition to having each GPT model generate and answer its own evaluation questions, we conduct a controlled experiment using Gemini 2.5 Pro as an independent question generator. This dual approach isolates pure evaluation behavior from question-generation capabilities, validating that the observed "evaluation personalities" are inherent model properties. Furthermore, our semantic similarity analysis of generated questions reveals systematic patterns in how different model families conceptualize quality assessment, with significant divergence between GPT and Gemini approaches despite consistent asymmetry between positive verification and negative detection strategies.

Rather than ranking model capabilities, this research uncovers how different AI architectures conceptualize and approach systematic evaluation, what aspects they prioritize, and what this reveals about the nature of AI assessment behavior. Our dual-assessment methodology provides a template for analyzing AI assessment patterns that captures both systematic precision and holistic understanding approaches, leveraging recent insights into evaluation granularity and vision-language model development~\citep{Lian2025DescribeAnything}.

 \section{Background}

\subsection{Large Language Models as Evaluators}

Evaluating vision-language outputs, such as image captions, has historically relied on
n-gram overlap metrics like BLEU~\citep{papineni2002bleu}, METEOR~\citep{banerjee2005meteor},
and ROUGE~\citep{lin2004rouge}, as well as vision-specific measures such as
SPICE~\citep{anderson2016spice} and CIDEr~\citep{vedantam2015cider}.
While effective for standardized benchmarking, these metrics often fail to
capture semantic nuances, coherence, and factual correctness, leading to
misalignment with human judgments. Embedding-based approaches, such as
Sentence-BERT~\citep{reimers2019sentence}, have improved correlation with human ratings
by measuring semantic similarity, but they remain limited in interpretability
and robustness.

Recently, \textit{Large Language Models as evaluators} have emerged as a
transformative paradigm. Instead of relying on rigid lexical overlap, LLMs
(e.g., GPT-4 and Claude) can be prompted to assess captions and responses
holistically. For instance, CLAIR~\citep{chan2023clair} employs GPT-4 in a
zero-shot setup to directly rate candidate captions against references,
achieving much higher correlation with human assessments compared to SPICE and
CLIP-based metrics. CLAIR not only provides scores but also textual
justifications, offering interpretability beyond traditional metrics.
Extensions such as CLAIR-A~\citep{wu2024clair} demonstrate similar gains
in audio captioning evaluation. More broadly, GPT-4V and Gemini
models~\citep{yang2023dawn,team2023gemini} have been shown to act as
reliable zero-shot judges in multimodal tasks, with consistency rivaling
human annotators.

This line of research has influenced training regimes as well: preference
optimization frameworks now incorporate LLM-based evaluation signals to
reward desirable model outputs~\citep{o1}. The trend toward
``evaluation by reasoning'' marks a shift from surface-level metrics to
context-aware, interpretable, and human-aligned assessment, positioning LLMs
as central evaluators in the multimodal era.

\subsection{Vision--Language Advances}

The vision--language modeling landscape has rapidly progressed from
task-specific architectures to general-purpose Multimodal Large Language Models (MLLMs).
Pioneering efforts such as Flamingo~\citep{alayrac2022flamingo} demonstrated the
effectiveness of coupling frozen language models with visual encoders for few-shot image understanding. BLIP~\citep{li2022blip} and BLIP-2~\citep{li2023blip}
further advanced this paradigm by efficiently aligning image and text
representations, enabling powerful pretraining with reduced trainable
parameters.

Instruction tuning has been key to unlocking interactive capabilities. LLaVA~\citep{li2024llavaonevisioneasyvisualtask} and its successors (LLaVA-Next~\citep{liu2024llavanext}, Visual Instruction Tuning~\citep{liu2023visual}) equip MLLMs with enhanced reasoning and OCR abilities. MiniGPT-4~\citep{chen2023minigpt} and ShareGPT4V~\citep{chen2023sharegpt4v} leverage GPT-4-quality captions to boost open-source models, while InstructBLIP~\citep{dai2023instructblip} introduces instruction-following at scale. Proprietary models such as GPT-4V~\citep{yang2023dawn},
Claude~\citep{team2024claude,team2025claude}, and Gemini~\citep{team2023gemini,team2024gemini} set new benchmarks across diverse multimodal tasks.

Beyond global image understanding, recent work emphasizes fine-grained and region-level reasoning. Grounding DINO~\citep{liu2024grounding} and Segment Anything (SAM)~\citep{kirillov2023segment,ravi2024sam} enable open-world detection and segmentation. Building on these, Segment-and-Caption Anything (SCA)~\citep{huang2024segment} attaches a lightweight captioning head to SAM, allowing region-level descriptions.
Ferret~\citep{you2023ferret,zhang2024ferret} extends this idea with granularity-aware grounding, handling points, boxes, and masks in natural
dialogue. Models like GLAMM~\citep{rasheed2024glamm} and OMG-LLaVA~\citep{zhang2024omg} aim to unify image, object, and pixel-level reasoning within a single framework. Furthermore, video-focused benchmarks such as VideoRefer~\citep{yuan2024videorefer} and systems like Elysium~\citep{wang2024elysium} extend multi-modal reasoning into the spatio-temporal domain.

Finally, flagship foundation models highlight the power of large-scale multimodal training Gemini~\citep{team2023gemini,team2024gemini}, Qwen-VL~\citep{bai2023qwen,wang2024qwen2,Qwen2.5-VL}, and PaLIX~\citep{chen2024scaling} show emergent capabilities in grounding, multilingual reasoning, and long-context video understanding. Together, these advances mark a decisive shift: vision--language research is moving toward unified foundation models that can perceive, reason, and act across modalities with human-like breadth.

\section{Methodology}

\subsection{Workflow Steps}

\begin{figure}[h!]
    \centering
    \includegraphics[width=\textwidth]{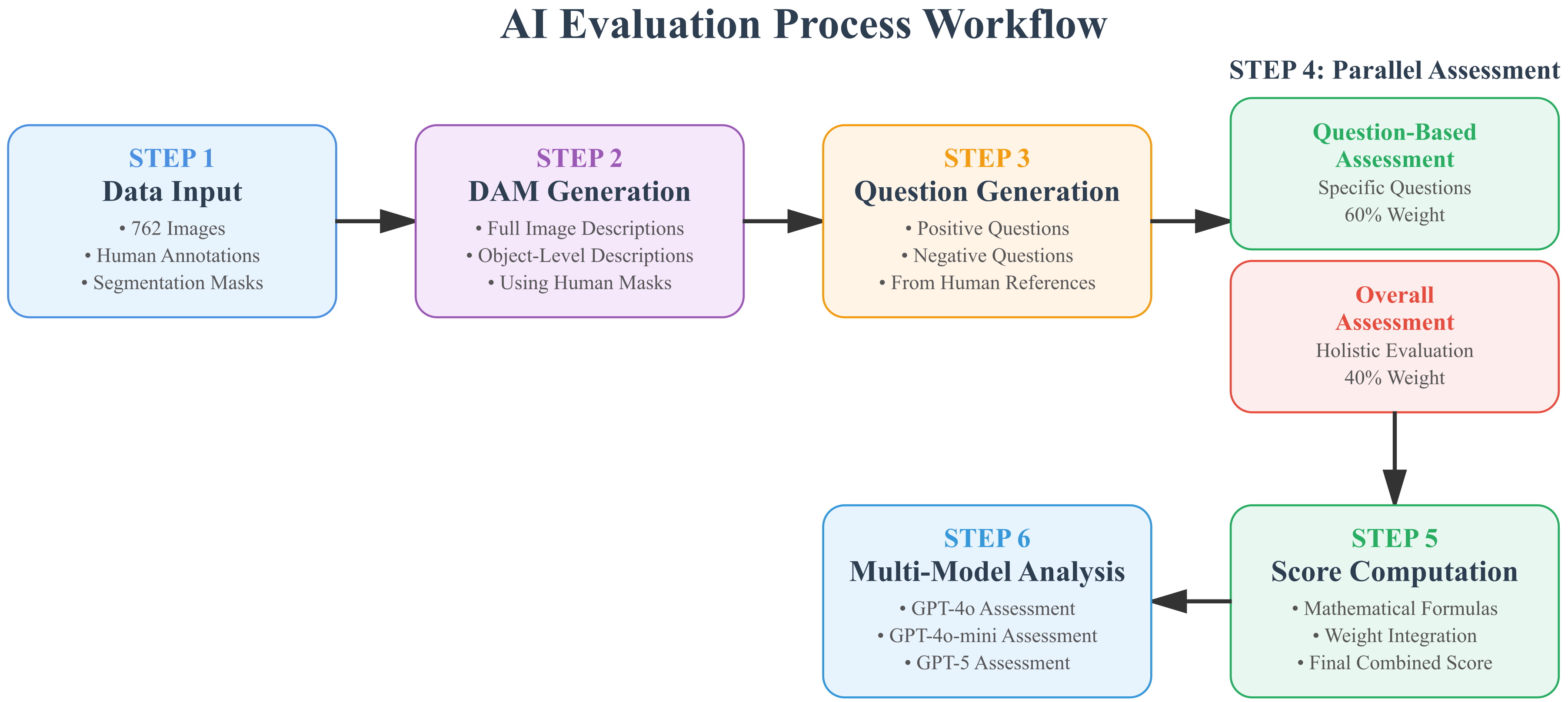}
    \caption{Complete evaluation methodology workflow. The pipeline progresses from dataset preparation through description generation, evaluation question formulation, dual assessment paradigms, scoring computation, and culminates in pattern analysis and bias identification across three GPT model variants.}
    \label{fig:workflow}
\end{figure}

Our experimental methodology employs a systematic multi-stage pipeline designed to comprehensively analyze evaluation patterns across different GPT architectures when assessing vision-language descriptions. The workflow, illustrated in Figure~\ref{fig:workflow}, consists of seven interconnected stages that progressively transform raw visual data into quantitative insights about AI evaluation behavior.

\subsubsection{Stage 1: Dataset Preparation and Preprocessing}
The initial stage involves the curation and preparation of image samples from the Dataset. Each sample undergoes preprocessing to extract human-annotated semantic segmentation masks, multi-tier textual descriptions (including titles, narrative descriptions, and technical scene analyses), and associated metadata. This stage ensures data consistency and establishes the ground truth baseline against which model-generated descriptions will be evaluated.

\subsubsection{Stage 2: Vision-Language Description Generation}
We leverage NVIDIA's Describe Anything Model (DAM-3B)~\citep{Lian2025DescribeAnything}, a vision-language model specifically designed for generating detailed descriptions of both full images and masked regions, to create rich textual annotations for our dataset. For each image in the DataSeeds validation set, we employ a two-stage description generation process: first, we generate a comprehensive scene-level description of the entire image focusing on environmental context, lighting, camera angles, and visual style (20-80 words); second, we iterate through each human-annotated segmentation mask, converting the polygon coordinates to binary masks and feeding them to DAM to generate object-specific descriptions. The model processes these inputs using a temperature of 0.2 and top-p of 0.5 to ensure consistent, focused outputs while maintaining descriptive quality. This approach allows us to augment the existing human annotations with AI-generated descriptions that capture both holistic scene understanding and fine-grained object details, creating a multi-level semantic representation where each segmented object receives its own contextual description within the broader scene narrative. The streaming generation pipeline processes the entire dataset with automatic checkpointing for robustness, ultimately producing a comprehensive JSON structure containing both scene-level and object-level descriptions paired with the original segmentation masks.

\subsubsection{Stage 3: Evaluation Question Synthesis}
The third stage involves the automatic generation of evaluation questions by each GPT judge (GPT-4o, GPT-4o-mini, and GPT-5). Using the human-generated ground truth descriptions as reference, each model independently synthesizes two categories of questions: (i) positive questions that verify the presence of specific attributes that should appear in accurate descriptions, and (ii) negative questions that check for the absence of incorrect or hallucinatory elements. This bidirectional questioning strategy enables comprehensive assessment of both descriptive accuracy and error avoidance.

\subsubsection{Stage 4: Dual Assessment Paradigm}
The fourth stage implements our dual assessment framework, executing two parallel evaluation streams. The question-based assessment utilizes the synthesized questions to systematically evaluate DAM descriptions through binary verification of specific attributes. Simultaneously, the holistic assessment evaluates three continuous dimensions: accuracy (factual correctness), completeness (coverage of key attributes), and hallucination penalty (presence of unsupported information). This parallel architecture captures both granular and gestalt aspects of description quality.

\subsubsection{Stage 5: Quantitative Score Computation}
In the fifth stage, we apply the mathematical formulations detailed in Section 2.2 to transform raw evaluation responses into standardized numerical scores. This includes computing weighted averages for positive and negative question responses, normalizing holistic assessment dimensions, and calculating the final combined score through our weighted integration formula. The standardization ensures comparability across different models and evaluation paradigms.

\subsubsection{Stage 6: Cross-Model Pattern Analysis}
The sixth stage performs comparative analysis across the three GPT models to identify distinctive evaluation patterns. This involves analysis of score distributions, and identification of model-specific emphasis areas. We employ visualization techniques including box plots, radar charts, and scatter plots to reveal assessment consistency, variance patterns, and inter-method dependencies.

This comprehensive workflow enables us to move beyond simple performance metrics to understand the underlying mechanisms and biases that govern how different AI architectures approach the complex task of evaluating vision-language descriptions.

\subsection{Use LLM as Judge}

Our evaluation methodology employs a sophisticated dual-assessment framework leveraging Large Language Models as autonomous judges to systematically evaluate the quality of vision-language descriptions generated by NVIDIA's DAM model. This framework combines granular attribute verification with holistic quality assessment, providing a comprehensive evaluation paradigm that captures both local and global aspects of descriptive accuracy. \footnote{For a detailed sample of how the LLMs are prompted to be used as a judge for such a task refer to Appendix \ref{APP_prompt_templates}}

\subsubsection{Holistic Evaluation}

The holistic evaluation component implements a three-dimensional assessment framework where LLM judges directly evaluate descriptions along orthogonal quality dimensions. We leverage the semantic understanding capabilities of LLMs to assess quality through natural language prompting. Each dimension is evaluated as a continuous score in the unit interval $[0,1]$, enabling nuanced quality quantification beyond binary classification.

\paragraph{Three-Dimensional Assessment Framework:}

Let ${D}$ represent the DAM-generated description and ${H}$ represent the human reference description. The LLM judge is prompted to evaluate three dimensions:

\begin{itemize}
    \item \textbf{Accuracy} $\in [0,1]$: \textit{Factual correctness of stated details}
    
    The LLM assesses how factually correct the details in the AI description are by evaluating:
    \begin{itemize}
        \item Whether colors, quantities, object types, and spatial relationships are correct
        \item The presence of factual errors or misidentifications
    \end{itemize}
    
    \item \textbf{Completeness} $\in [0,1]$: \textit{Coverage of key reference attributes}
    
    The LLM measures how comprehensively the AI description covers the human references by examining:
    \begin{itemize}
        \item Whether main subjects and key details are mentioned
        \item Whether important technical aspects (camera angle, mood, setting) are captured
    \end{itemize}
    
    \item \textbf{Hallucination Penalty} $\in [0,1]$: \textit{Incorrect or unsupported information penalty}
    
    The LLM quantifies the presence of unsupported information by identifying:
    \begin{itemize}
        \item Details added that are not mentioned in human descriptions
        \item Claims that contradict the human references
    \end{itemize}
\end{itemize}

\paragraph{LLM Prompting Strategy:}

Each LLM judge receives a structured prompt requesting evaluation scores for the three dimensions. The prompt explicitly defines the evaluation criteria and score ranges, ensuring consistent assessment across samples. Formally, the LLM evaluation function can be represented as:

\begin{equation}
{E}_{\text{LLM}}: ({D}, {H}) \rightarrow \{{A}, {C}, {P}\}
\end{equation}

where ${A}$, ${C}$, and ${P}$ represent the accuracy, completeness, and hallucination penalty scores respectively, each constrained to $[0,1]$.

\paragraph{Overall Holistic Score Formulation:}

Given the three dimension scores produced by the LLM judge, the overall holistic assessment score $S_{\text{holistic}}$ is computed as a weighted linear combination with penalty adjustment:

\begin{equation}
S_{\text{holistic}} = \omega_{A} \cdot {A} + \omega_{C} \cdot {C} - \omega_{P} \cdot {P}
\end{equation}

where:
\begin{itemize}
    \item $\omega_{A} = 0.6$ (accuracy weight)
    \item $\omega_{C} = 0.4$ (completeness weight)
    \item $\omega_{P} = 1.0$ (hallucination penalty weight)
\end{itemize}

The final score is normalized to the unit interval through clipping:

\begin{equation}
\hat{S}_{\text{holistic}} = \text{clip}_{[0,1]}(S_{\text{holistic}}) = \max\{0, \min\{1, S_{\text{holistic}}\}\}
\end{equation}

This formulation prioritizes factual accuracy while maintaining substantial weight on completeness, and applies a full penalty for hallucinated content to discourage the generation of unsupported information.

\subsubsection{Question-Based Evaluation}

The question-based evaluation paradigm, inspired by the DLC-Bench methodology \citep{Lian2025DescribeAnything}, operationalizes assessment through targeted verification questions. This approach transforms the evaluation task into a series of binary verification problems, enabling precise quantification of specific descriptive attributes.

\paragraph{Question Generation Framework:}

Let ${Q} = {Q}^+ \cup {Q}^-$ represent the complete set of evaluation questions generated by the LLM judge, where:

\begin{itemize}
    \item ${Q}^+ = \{q_1^+, q_2^+, \ldots, q_{n^+}^+\}$ denotes positive verification questions
    \item ${Q}^- = \{q_1^-, q_2^-, \ldots, q_{n^-}^-\}$ denotes negative exclusion questions
\end{itemize}

Each question $q_i$ is formulated to test one specific factual detail that can be objectively verified. The LLM generates these questions using exact words and phrases from human descriptions to ensure precision.

\paragraph{Scoring Functions:}

\textbf{Positive Question Scoring:}

For each positive question $q_i^+ \in {Q}^+$, we define the scoring function:

\begin{equation}
\sigma(q_i^+, {D}) = \begin{cases}
1 & \text{if attribute is correctly and clearly mentioned} \\
0 & \text{if attribute is omitted} \\
-0.5 & \text{if attribute is mentioned but factually incorrect}
\end{cases}
\end{equation}

The aggregate positive score is computed as:

\begin{equation}
S^+ = \frac{1}{|{Q}^+|} \sum_{i=1}^{|{Q}^+|} \sigma(q_i^+, {D})
\end{equation}

\textbf{Negative Question Scoring:}

For each negative question $q_i^- \in {Q}^-$, the scoring function is:

\begin{equation}
\sigma(q_i^-, {D}) = \begin{cases}
1 & \text{if incorrect detail is correctly avoided} \\
-1 & \text{if incorrect detail is erroneously included}
\end{cases}
\end{equation}

The aggregate negative score is:

\begin{equation}
S^- = \frac{1}{|{Q}^-|} \sum_{i=1}^{|{Q}^-|} \sigma(q_i^-, {D})
\end{equation}

\paragraph{Normalized Question-Based Score:}

The scores are normalized to ensure non-negativity:

\begin{align}
\hat{S}^+ &= \max\{0, S^+\} \\
\hat{S}^- &= \max\{0, S^-\}
\end{align}

The combined question-based score is:

\begin{equation}
S_{\text{question}} = \alpha \cdot \hat{S}^+ + (1-\alpha) \cdot \hat{S}^-
\end{equation}

where $\alpha = 0.5$ represents equal weighting between positive and negative assessments.

\paragraph{Integrated Evaluation Score:}

The final integrated evaluation score $S_{\text{final}}$ combines both assessment paradigms:

\begin{equation}
S_{\text{final}} = \beta \cdot S_{\text{question}} + (1-\beta) \cdot \hat{S}_{\text{holistic}}
\end{equation}

where $\beta = 0.6$ prioritizes question-based assessment while maintaining holistic evaluation influence. The final score is constrained to the unit interval:

\begin{equation}
\hat{S}_{\text{final}} = \text{clip}_{[0,1]}(S_{\text{final}}) = \max\{0, \min\{1, S_{\text{final}}\}\}
\end{equation}

\subsubsection{GPT Model Architectures as Judges}

We employ three distinct GPT architectures as evaluation judges, each representing different evolutionary stages in large language model development. This multi-model approach enables us to identify both universal evaluation patterns and architecture-specific biases.

\paragraph{GPT-4o (Baseline Architecture)~\citep{openai2024gpt4ocard}:} 

GPT-4o represents the established GPT-4 family architecture, serving as our baseline evaluation model. This model employs a transformer-based architecture with demonstrated capabilities across diverse language understanding tasks.

\paragraph{GPT-4o-mini (Efficient Variant)~\citep{openai2024gpt4ocard}:} 

GPT-4o-mini implements an architecturally optimized variant of GPT-4o, designed for computational efficiency while preserving evaluation quality. As we will discuss later, despite parameter reduction, our empirical analysis reveals that GPT-4o-mini exhibits superior systematic consistency in assessment tasks. This consistency suggests potential benefits of architectural streamlining for evaluation tasks.

\paragraph{GPT-5 (Next-Generation Architecture)~\citep{openai2025gpt5}:} 

Released on August 7, 2025, GPT-5 represents a paradigmatic shift in model architecture. The model implements a unified framework with dynamic routing capabilities that automatically modulate computational depth based on query complexity. This architecture achieves state-of-the-art performance on multiple benchmarks:

\begin{table}[h!]
\centering
\caption{GPT-5 Performance Benchmarks}
\begin{tabular}{lc}
\toprule
\textbf{Benchmark} & \textbf{Performance} \\
\midrule
AIME 2025 & 94.6\% \\
SWE-bench Verified & 74.9\% \\
MMMU (Multimodal) & 84.2\% \\
Factual Error Reduction & 80\% (vs GPT-4) \\
\bottomrule
\end{tabular}
\label{tab:gpt5_benchmarks}
\end{table}

\subsection{Question Similarity Analysis}

In order to have a better understanding on the generated question from the LLM judge models we proposed a systematic analysis on the similarity of the generated questions in the embedding space. We investigated the similarity of evaluation questions generated by three large language models across the dataset samples. Using semantic embedding analysis and cosine similarity measures, we examine how consistently different LLMs formulate positive and negative evaluation questions. Our findings reveal interesting differences in questioning strategies between positive and negative question types, with important implications for cross-model evaluation reliability. 

We assess the whole of samples individually to ensure robust statistical analysis. As it is shown in Algorithm \ref{alg:similarity}, For every sample, we extract the positive and negative questions generated by each LLM pair, then converts these questions into high-dimensional semantic embeddings using Sentence-BERT~\citep{reimers-2019-sentence-bert}. It constructs a cosine similarity matrix comparing all question pairs between the two LLMs, then computes the symmetric maximum similarity by averaging the best matches in both directions - this metric captures how well each question from one LLM is represented by the most similar question from another LLM. The algorithm repeats this process across all samples and question types, finally aggregating the results to produce overall similarity scores that reflect consistent patterns while accounting for sample-level variation. This approach ensures that the final similarity measures represent genuine cross-model consistency rather than isolated agreements on specific samples.

\begin{algorithm}
\caption{Question similarity analysis}
\label{alg:similarity}
\begin{algorithmic}[1]
\ForAll{sample $k$ in $N$ samples}
    \ForAll{LLM pair $(i, j)$}
        \State Extract $\text{positive\_questions}_i[k]$ and $\text{positive\_questions}_j[k]$
        \State Extract $\text{negative\_questions}_i[k]$ and $\text{negative\_questions}_j[k]$
        \State
        \Comment{Compute embeddings using Sentence-BERT}
        \State $\text{embeddings}_i \gets \text{encode}(\text{questions}_i)$
        \State $\text{embeddings}_j \gets \text{encode}(\text{questions}_j)$
        \State
        \Comment{Calculate cosine similarity matrix}
        \State $\text{similarity\_matrix} \gets \text{cosine\_similarity}(\text{embeddings}_i, \text{embeddings}_j)$
        \State
        \Comment{Symmetric maximum similarity}
        \State $\text{max}_{i \to j} \gets \text{average}(\text{max\_along\_rows}(\text{similarity\_matrix}))$
        \State $\text{max}_{j \to i} \gets \text{average}(\text{max\_along\_columns}(\text{similarity\_matrix}))$
        \State $\text{symmetric\_similarity} \gets (\text{max}_{i \to j} + \text{max}_{j \to i}) / 2$
        \State
        \State $\text{store\_result}(\text{sample}_k, \text{pair}_{ij}, \text{symmetric\_similarity})$
    \EndFor
\EndFor
\State
\Comment{Aggregate across all samples}
\State $\text{final\_similarity} \gets \text{average\_across\_samples}(\text{symmetric\_similarity})$
\end{algorithmic}
\end{algorithm}

\subsection{Dataset}
For our experiments, we use the entire validation set of the DataSeeds.AI (DSD) dataset \citep{abdoli2025peer} since it has rich annotations from human experts. Each sample in the dataset includes comprehensive multi-tier annotations quite suitable for this study:

\begin{itemize}
    \item Concise descriptive titles encapsulating the central subjects of each image
    \item Detailed narrative descriptions, systematically capturing object identifiers, spatial relationships, and objective documentation of visual elements
    \item Technical scene analyses documenting photographic methodologies, camera angles (eye-level, elevated, low, or bird's-eye), photographic style, and textual elements
    \item Full semantic segmentation masks defining precise pixel-level contours of objects, capturing their true morphology including occluded and overlapping forms
\end{itemize}

As noted by \citep{abdoli2025peer}, this dataset uniquely integrates professional photographic standards with semi-professional creative expression, presenting a wide-ranging visual dataset with inherent hierarchical quality signals derived from GuruShots' community-driven ranking system. Each image inherently carries an evaluative dimension based on human preferences and perceptions.

Figure~\ref{fig:sample} shows a representative sample from the DSD Dataset demonstrating the structured technical annotation designed to capture salient features of photographic practice. The inference of DAM model for this example is also shown in Appendix \ref{APP_DAM_inf_sample}  

\begin{figure}[h!]
    \centering
    \includegraphics[width=0.8\textwidth]{files/image3.jpg}
    \caption{Sample from DataSeeds.AI Dataset showing image titles, human descriptions, and segmented objects. The image shows three long-tailed Tits standing on metal structures against a blurred background.}
    \label{fig:sample}
\end{figure}

\begin{itemize}
    \item \textit{\textbf{Title:}} Feathered Trio
    \item \textit{\textbf{Image description:}} A picture of three long-tailed Tits standing on metal structures against a blurred background highlighting the colorful feathers of these cute little creatures.
    \item \textit{\textbf{Scene description:}} A close up picture of three long-tailed tits in natural, soft lighting. The atmosphere feels cute and soft, enhanced by a muted urban color palette of pinks, whites, grays, and black against a green and white blurred background.
\end{itemize}

The use of semantic segmentation masks, as emphasized in the DSD paper, was crucial for our evaluation. Unlike bounding boxes which delineate crude rectangular regions prone to capturing irrelevant background artifacts, semantic segmentation defines the precise pixel-level contours of objects. This granularity yields more accurate scene understanding and enables finer-grained feature extraction, allowing models to infer depth, hierarchy, and adjacency relationships. For our study, these human-annotated segmentation masks ensured that DAM's object-level descriptions corresponded exactly to the regions identified by human evaluators, providing a controlled foundation for subsequent GPT model assessment.

The DSD's multi-tier annotation strategy, which closely resembles NVIDIA's Describe Anything Model (DAM) framework and its semi-supervised data pipeline, combines structured human annotations with scalable machine-assisted processes to enhance data quality and efficiency. This procedural rigor, with each mask subject to manual review and verification by trained annotators, ensures the dataset's reliability for consistent benchmarking and reproducible improvements in model evaluation.

\section{Results}

Our comprehensive analysis of 762 assessments, using the DSD validation set, across three models revealed statistically significant differences in evaluation approaches, with distinct patterns emerging in how each model conceptualizes and executes assessment tasks. The following sections present a detailed examination of model performance metrics, evaluation personalities, bias patterns, and visual analysis insights that collectively reveal fundamental characteristics of AI evaluation behavior.

\subsection{Model Assessment Performance}\label{model_assessmet_per}

Table~\ref{tab:assessment_scores} presents the comprehensive assessment scores across all evaluation dimensions for the three GPT models evaluated on 762 samples. The results reveal substantial heterogeneity in both absolute performance levels and variance patterns across models, suggesting fundamentally different approaches to evaluation tasks. For a detailed example of the analysis for the sample shown in Figure \ref{fig:sample}, refer to Appendix \ref{APP_Detailed_AI judge_sample}.

\begin{table}[h!]
\centering
\caption{Assessment Scores (762 samples each). Bold values indicate the highest score in each assessment dimension. For all metrics except Hallucination Vigilance, higher means the LLM judge scores the DAM model description higher. For Hallucination Vigilance, higher score means the model adds more hallucination penalty.}
\begin{tabular}{lccc}
\toprule
\textbf{Assessment Dimension} & \textbf{GPT-4o} & \textbf{GPT-4o-mini} & \textbf{GPT-5} \\
\midrule
Question-Based Assessment & 67.10\% ($\pm$12.70\%) & \textbf{68.79\% ($\pm$11.99\%)} & 59.00\% ($\pm$13.23\%) \\
Overall Assessment & 58.56\% ($\pm$11.40\%) & \textbf{64.02\% ($\pm$0.43\%)} & 17.82\% ($\pm$19.98\%) \\
Combined Assessment & 63.68\% ($\pm$9.74\%) & \textbf{66.88\% ($\pm$7.19\%)} & 42.53\% ($\pm$12.91\%) \\
Accuracy Emphasis & 81.82\% ($\pm$6.29\%) & \textbf{85.01\% ($\pm$0.18\%)} & 61.08\% ($\pm$15.27\%) \\
Completeness Focus & 69.96\% ($\pm$5.45\%) & \textbf{70.01\% ($\pm$0.36\%)} & 64.48\% ($\pm$15.35\%) \\
Hallucination Vigilance & 18.55\% ($\pm$6.43\%) & 14.99\% ($\pm$0.18\%) & \textbf{54.10\% ($\pm$18.84\%)} \\
Positive Confirmation & 41.98\% ($\pm$18.28\%) & \textbf{45.47\% ($\pm$17.40\%)} & 38.82\% ($\pm$17.97\%) \\
Negative Detection & \textbf{92.21\% ($\pm$14.29\%)} & 92.12\% ($\pm$14.95\%) & 79.17\% ($\pm$26.58\%) \\
\bottomrule
\end{tabular}
\label{tab:assessment_scores}
\end{table}

The question-based assessment reveals that GPT-4o-mini assigns the highest score to DAM’s inference performance at 68.79\%, marginally exceeding the 67.10\% assigned by GPT-4o. However, the most striking pattern emerges not from the mean scores but from the variance characteristics. GPT-4o-mini demonstrates a remarkably low standard deviation ($\pm$11.99\%), suggesting consistent application of evaluation criteria across diverse samples. In contrast, GPT-5 exhibits the highest variance ($\pm$13.23\%), indicating substantial context-dependent variation in its assessment approach.

The overall assessment dimension presents the most dramatic divergence among evaluators. GPT-4o-mini assigns DAM the highest score at 64.02\%, but with an extraordinarily low standard deviation of only $\pm$0.43\%. This near-zero variance represents an unprecedented level of consistency in holistic evaluation, suggesting that GPT-4o-mini applies virtually identical assessment criteria regardless of content variation. GPT-4o assigns DAM a moderate score of 58.56\% with reasonable variance ($\pm$11.40\%), while GPT-5 assigns an extremely low score of 17.82\% with very high variance ($\pm$19.98\%). This pattern suggests that GPT-5's architectural innovations, while potentially beneficial for general tasks, may introduce instability in structured evaluation contexts.

The combined assessment, which integrates both question-based and holistic evaluations with a 60:40 weighting, shows GPT-4o-mini again assigning the highest score to DAM at 66.88\% with controlled variance ($\pm$7.19\%). GPT-4o follows at 63.68\% ($\pm$9.74\%), while GPT-5 recovers somewhat to 42.53\% ($\pm$12.91\%), benefiting from the stabilizing influence of the question-based component.

Examining the individual assessment components reveals nuanced patterns. In the accuracy emphasis dimension, GPT-4o-mini assigns DAM 85.01\% with near-perfect consistency ($\pm$0.18\%), while GPT-4o assigns 81.82\% ($\pm$6.29\%). GPT-5’s accuracy assessment drops to 61.08\% with high variance ($\pm$15.27\%), suggesting difficulty in maintaining consistent accuracy standards. The completeness focus dimension shows similar trends, with GPT-4o-mini assigning 70.01\% ($\pm$0.36\%) and GPT-4o assigning 69.96\% ($\pm$5.45\%), both comparable, while GPT-5 assigns 64.48\% but with substantial variance ($\pm$15.35\%).

The hallucination vigilance metric reveals a dramatic reversal of patterns. GPT-5 assigns DAM a very high penalty rate of 54.10\%, indicating hypersensitivity to potential hallucinations. This is accompanied by the highest variance ($\pm$18.84\%) among all dimensions, reflecting inconsistent application of hallucination criteria. GPT-4o and GPT-4o-mini assign more moderate hallucination penalties at 18.55\% and 14.99\% respectively, with GPT-4o-mini again demonstrating exceptional consistency ($\pm$0.18\%).

The positive confirmation and negative detection dimensions show a universal pattern across evaluators. Positive confirmation rates remain uniformly low, with GPT-4o-mini assigning DAM the highest score at 45.47\%, followed by GPT-4o at 41.98\% and GPT-5 at 38.82\%. All models show high variance in this dimension (17-–18\%), suggesting inherent difficulty in confirming correct information. Conversely, negative detection shows uniformly high scores, with GPT-4o assigning 92.21\% and GPT-4o-mini 92.12\%, both with moderate variance (14--15\%). GPT-5 assigns a lower score of 79.17\% with extreme variance ($\pm$26.58\%), indicating inconsistent error detection capabilities.

\subsection{Model Evaluation Personalities}\label{model_personalities}
The quantitative patterns observed in Table~\ref{tab:assessment_scores} coalesce into three distinct evaluation personalities of the GPT judges, each representing a fundamentally different approach to scoring DAM’s outputs. These personalities emerge not from isolated metrics but from the complex interplay of assigned score levels, variance patterns, and dimensional emphasis.

\paragraph{GPT-4o-mini: The Systematic Consistency Assessor}

GPT-4o-mini emerges as the paradigm of systematic judgment, characterized by exceptional consistency across all assessment dimensions. Its variance patterns are unprecedented in evaluation literature, with standard deviations ranging from $\pm$0.18\% to $\pm$17.40\%, and, critically, achieving near-zero variance ($\pm$0.43\%) in overall assessment. This consistency extends beyond simple reliability; it represents a fundamental approach to judging that prioritizes systematic application of fixed criteria over context-dependent adaptation.

As a judge, GPT-4o-mini assigns DAM the highest scores in seven of eight assessment dimensions, demonstrating not only consistency but also generosity in its evaluations. The 85.01\% accuracy emphasis with $\pm$0.18\% variance suggests that GPT-4o-mini applies a stable notion of accuracy criteria across diverse content. Similarly, its 70.01\% completeness focus with $\pm$0.36\% variance indicates systematic coverage assessment independent of specific domain characteristics.

GPT-4o-mini’s lowest hallucination penalty rate (14.99\% $\pm$0.18\%) suggests a balanced approach to detecting false information, avoiding both excessive skepticism and insufficient vigilance. This model’s judging philosophy can be characterized as: Apply consistent, systematic criteria across all evaluation dimensions representing an algorithmic approach to assessment that prioritizes reproducibility and fairness over nuanced adaptation.

\paragraph{GPT-4o: The Specialized Error Detector}

GPT-4o presents a more specialized judging profile, characterized by exceptional negative detection capability (92.21\% $\pm$14.29\%) while maintaining balanced performance across other dimensions. This specialization suggests a judging bias toward error identification, making GPT-4o particularly effective in quality control contexts.

Its moderate variance patterns ($\pm$5.45\% to $\pm$18.28\%) indicate controlled adaptation to content variation without the extreme rigidity of GPT-4o-mini or the instability of GPT-5. The 81.82\% accuracy emphasis and 69.96\% completeness focus reflect strong evaluation capacity, while the 18.55\% hallucination penalty demonstrates a balanced approach to false information detection.

GPT-4o’s judging philosophy can be summarized as: Excel at identifying what’s wrong while maintaining overall assessment quality. This makes it well-suited for scenarios where error detection is paramount, such as fact-checking or quality assurance. Its ability to maintain high negative detection rates while achieving reasonable positive confirmation (41.98\%) indicates sophisticated discrimination rather than simple negative bias.

\paragraph{GPT-5: The Inconsistent High-Threshold Assessor}

GPT-5 exhibits the most complex and problematic judging profile. Its extreme hallucination vigilance (54.10\% $\pm$18.84\%) coupled with very low overall assessment scores for DAM (17.82\% $\pm$19.98\%) suggests a fundamental mismatch between its architectural innovations and the requirements of consistent evaluation.

Its variance patterns are the highest across dimensions, ranging from $\pm$12.91\% to $\pm$26.58\%, indicating substantial instability in how it applies judging criteria. While GPT-5 assigns DAM a 61.08\% score in accuracy emphasis, this marks a notable decline compared to the GPT-4 family. Its 79.17\% negative detection rate, though respectable, falls substantially below its predecessors.

GPT-5’s judging philosophy can be summarized as: \textit{Apply highly variable standards with extreme conservatism toward potential false information.”} This pattern suggests that optimizations for broader intelligence tasks may introduce instability in evaluative contexts. GPT-5 appears to shift its judging criteria depending on context, reflecting more sophisticated contextual awareness that paradoxically reduces consistency and reliability in DAM evaluation.

\subsection{Universal Assessment Bias Discovery}
Analysis of the assessment patterns reveals systematic biases in how GPT judges evaluate DAM, suggesting fundamental characteristics of current AI evaluation approaches. These biases manifest most clearly in the asymmetry between positive confirmation and negative detection.

The negative detection dimension demonstrates remarkable consistency across the GPT-4 family, with GPT-4o assigning DAM 92.21\% ($\pm$14.29\%) and GPT-4o-mini assigning 92.12\% ($\pm$14.95\%) in error detection tasks. Even GPT-5, despite its overall instability, assigns a comparatively strong 79.17\% with substantially higher variance ($\pm$26.58\%). This universal proficiency in detecting incorrect information suggests that current training methodologies and architectures are optimized for error detection, likely reflecting the emphasis on minimizing harmful or factually incorrect outputs in AI safety training.

By contrast, positive confirmation emerges as a universal weakness across all evaluators. Across GPT judges, DAM receives positive confirmation rates below 50\%, with GPT-4o-mini assigning the highest at only 45.47\% ($\pm$17.40\%), followed by GPT-4o at 41.98\% ($\pm$18.28\%) and GPT-5 at 38.82\% ($\pm$17.97\%). This combination of consistently low mean scores and high variance (17–18\% across evaluators) suggests that confirming correct information presents inherent challenges that current architectures have not resolved.

The magnitude of this asymmetry is striking. GPT-4o shows a 2.20:1 ratio of negative detection to positive confirmation (92.21\%:41.98\%), GPT-4o-mini shows 2.03:1 (92.12\%:45.47\%), and GPT-5 shows 2.04:1 (79.17\%:38.82\%). This consistent 2:1 to 2.2:1 bias toward negative detection indicates a fundamental pattern in AI evaluation behavior that appears independent of specific model architectures.

This asymmetry has profound implications for evaluation design. Current GPT judges are inherently biased toward criticism over confirmation, which may lead to systematically deflated scores when used for quality assessment. The pattern suggests that training data and optimization objectives emphasize error avoidance over positive validation, creating evaluators that function more as critics than balanced assessors.

The consistency of this bias across GPT architectures suggests it may stem from fundamental aspects of training paradigms rather than specific model designs. Possible explanations include training data distributions that overrepresent error correction, reinforcement structures that penalize false positives more heavily than false negatives, or inherent asymmetries in how language models process negative versus positive statements.

\subsection{Visual Analysis Insights}

The visual analysis provides crucial insights into the underlying mechanisms driving the observed evaluation patterns. Four key visualizations reveal complementary aspects of model behavior that collectively illuminate the nature of AI evaluation approaches.
\begin{figure}[h!]
    \centering
    \includegraphics[width=\textwidth]{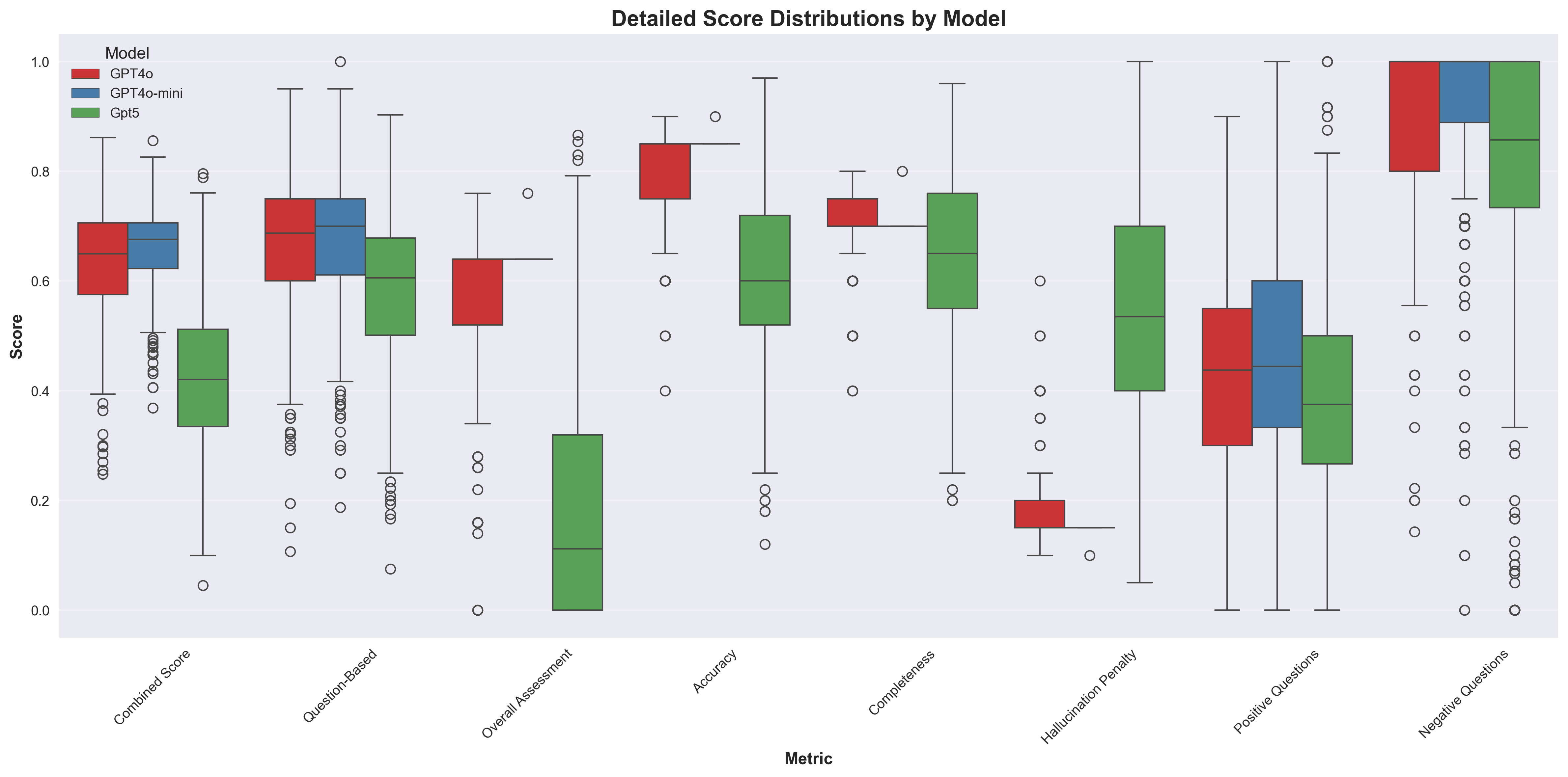}
    \caption{Score Distribution Patterns Reveal Model Assessment Consistency and Variability. The box plots illustrate the distribution of scores across eight assessment dimensions for each model. Box boundaries represent quartiles, whiskers extend to 1.5 times the interquartile range, and circles indicate outliers. The width of distributions directly reflects evaluation consistency, with narrower boxes indicating more systematic assessment approaches.}
    \label{fig:score_distribution}
\end{figure}

\paragraph{Distribution Analysis (Figure~\ref{fig:score_distribution})}

The score distribution patterns presented in Figure~\ref{fig:score_distribution} provide a comprehensive visualization of assessment consistency across models and dimensions. The most striking feature is GPT-4o-mini's extraordinary consistency in the Overall Assessment dimension, where the distribution collapses to essentially a horizontal line at approximately 0.64, with a standard deviation of merely $\pm$0.0043. This near-zero variance is unprecedented in evaluation literature and suggests that GPT-4o-mini applies virtually identical holistic assessment criteria regardless of content variation. The absence of visible box structure in this dimension indicates that the interquartile range approaches zero, meaning that 50\% of all assessments fall within an extremely narrow score band.

In contrast, GPT-5 demonstrates massive distribution spread in the Overall Assessment dimension, with scores ranging from near 0\% to above 80\% and numerous outliers at both extremes. The wide box structure and extended whiskers indicate that GPT-5's holistic evaluation varies dramatically across samples. The presence of outliers at both extremes suggests bimodal or multimodal evaluation behavior, where the model may be applying different evaluation frameworks to different content categories.

GPT-4o exhibits moderate distributions across most dimensions, with controlled variance that suggests balanced evaluation approaches. The model's distributions show regular box structures with symmetric whiskers, indicating approximately normal score distributions without the extreme consistency of GPT-4o-mini or the instability of GPT-5.

A universal pattern emerges in the comparison between Positive Confirmation and Negative Detection dimensions. All models show compressed distributions for Negative Detection, with scores concentrated in the 0.8-1.0 range and relatively narrow boxes. This compression indicates consistent high performance in error detection across diverse content. Conversely, Positive Confirmation distributions span the entire 0.0-1.0 range with wide boxes and extended whiskers, revealing substantial variability in the ability to confirm correct information.

The Hallucination Penalty dimension reveals distinct model philosophies. GPT-4o-mini and GPT-4o show narrow distributions concentrated near 0.15-0.20, indicating consistent, moderate hallucination detection. GPT-5's distribution in this dimension is dramatically different, with a wide box centered around 0.54 and extensive whiskers reaching both extremes. The numerous outliers in GPT-5's hallucination assessment suggest that the model alternates between extreme skepticism and minimal hallucination detection depending on content characteristics.

\begin{figure}[h!]
    \centering
    \includegraphics[width=0.75\textwidth]{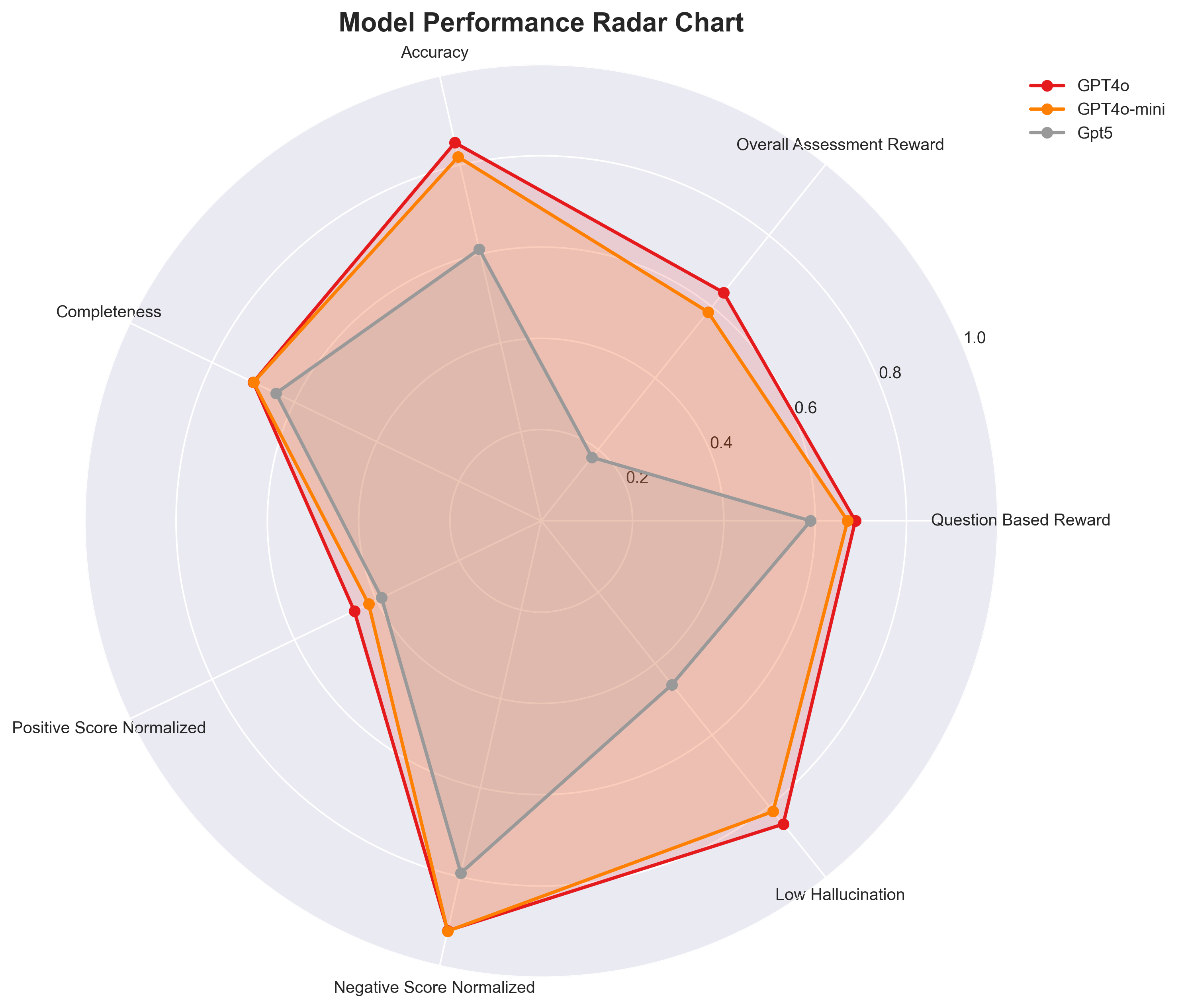}
    \caption{Radar Chart Reveals Distinct Model Assessment Emphasis Patterns. Each vertex represents a normalized assessment dimension, with distance from center indicating performance level. The polygon area reflects overall evaluation capability, while shape characteristics indicate dimensional emphasis. "Low Hallucination" is computed as (1 - Hallucination Penalty) to maintain consistent interpretation where larger values indicate better performance.}
    \label{fig:radar_chart}
\end{figure}

\paragraph{Assessment Emphasis Patterns (Figure~\ref{fig:radar_chart})}

The radar chart visualization in Figure~\ref{fig:radar_chart} synthesizes multidimensional performance into interpretable geometric patterns. GPT-4o-mini (orange polygon) presents the largest and most regular hexagonal shape, indicating balanced high performance across all dimensions. The polygon's vertices reach approximately 0.85 for Accuracy, 0.70 for Completeness, and maintain consistent distances from the center across other dimensions. This regular shape suggests systematic evaluation without dimensional bias.

GPT-4o (red polygon) shows a slightly smaller but still well-formed shape with a notable extension toward Negative Score Normalized, where it marginally exceeds GPT-4o-mini. This protrusion reflects GPT-4o's specialization in error detection while maintaining reasonable performance across other dimensions. The polygon's general regularity with this single emphasis supports the model's characterization as a specialized error detector with balanced general capabilities.

GPT-5 (gray polygon) presents a dramatically contracted and irregular shape, with the polygon area approximately 40\% of GPT-4o-mini's coverage. The shape shows severe contractions in Accuracy (0.61), Overall Assessment Reward (0.18), and Question Based Reward (0.59) dimensions. The only dimension where GPT-5 approaches the other models is Completeness (0.64), suggesting that while the model can identify coverage of topics, it struggles with accuracy assessment and overall quality evaluation. The irregular, contracted shape visually represents GPT-5's inconsistent and generally lower evaluation performance.

The "Low Hallucination" dimension, computed as the inverse of hallucination penalty for consistent interpretation, shows GPT-4o-mini and GPT-4o with similar extensions (approximately 0.85 and 0.81 respectively), while GPT-5 shows dramatic contraction (0.46). This visual representation emphasizes GPT-5's extreme conservatism regarding potential hallucinations, which may contribute to its overall lower assessment scores.

\begin{figure}[h!]
    \centering
    \includegraphics[width=0.75\textwidth]{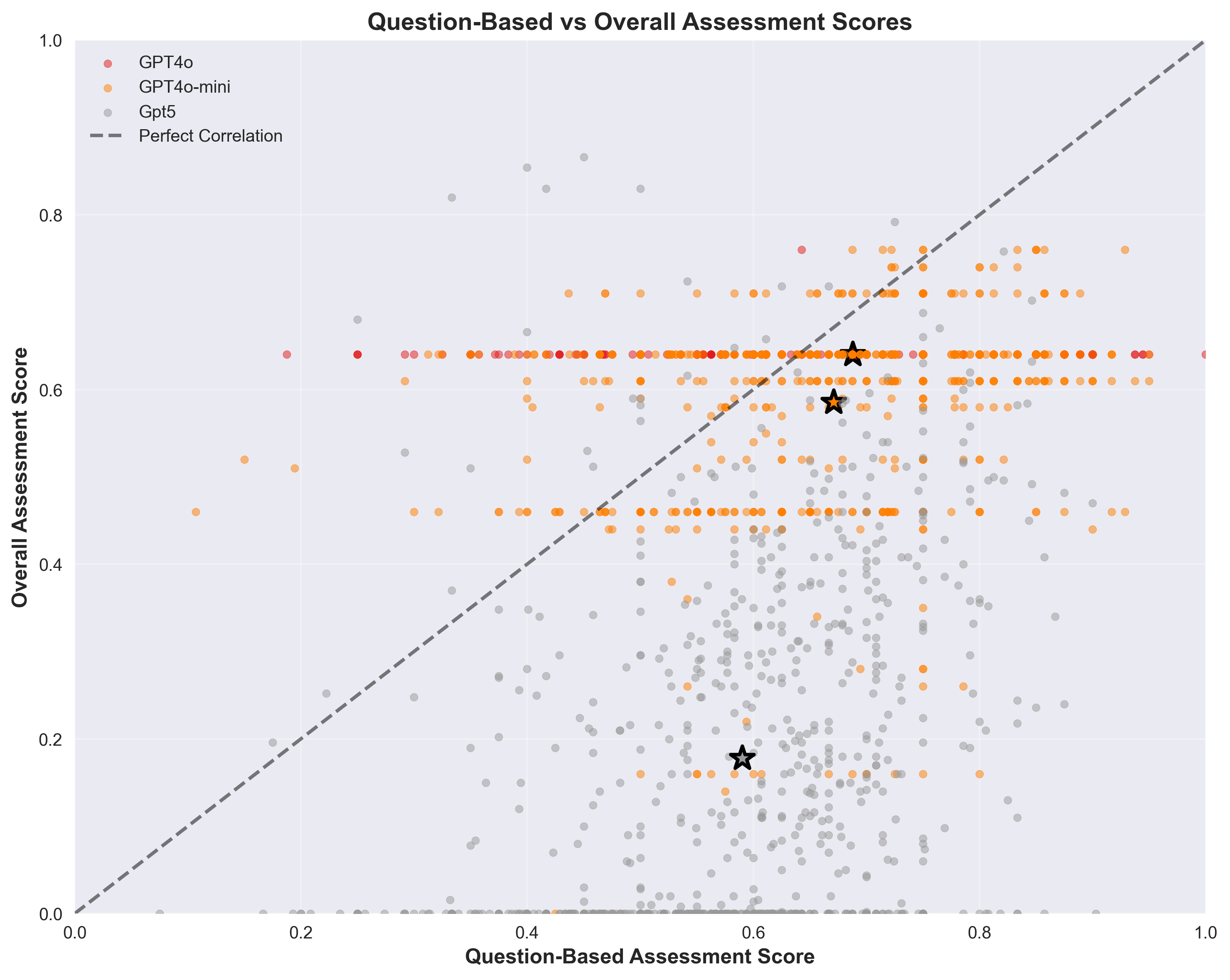}
    \caption{Assessment Method Integration Reveals Different Evaluation Strategies. The scatter plot displays the relationship between question-based (x-axis) and overall assessment (y-axis) scores for each model. Point density indicates score frequency, while spatial patterns reveal integration strategies. The diagonal reference line represents perfect correlation between assessment methods. Stars indicate mean positions for each model.}
    \label{fig:scatter_plot}
\end{figure}

\paragraph{Assessment Method Integration (Figure~\ref{fig:scatter_plot})}

Figure~\ref{fig:scatter_plot} reveals fundamental differences in how models integrate structured question-based assessment with holistic evaluation. The most remarkable pattern is GPT-4o-mini's (orange points) formation of a dense horizontal band at y=0.64, creating an almost perfect horizontal line across the full range of question-based scores (0.2 to 1.0). This extraordinary pattern, with a correlation coefficient of -0.014, indicates complete independence between assessment methods. GPT-4o-mini appears to compute holistic assessment using entirely separate mechanisms from question-based evaluation, treating them as orthogonal dimensions rather than related measures of quality.

GPT-4o (red points) demonstrates weak positive correlation (r=0.231) with moderate scatter around a generally positive trend. The points show increasing overall assessment scores with higher question-based scores, though with substantial variance. This pattern suggests that GPT-4o partially integrates information from structured questions into its holistic assessment while maintaining independent evaluation components. The scatter pattern forms a diffuse cloud with a slight upward trend, indicating probabilistic rather than deterministic integration of assessment methods.

GPT-5 (gray points) exhibits the highest correlation (r=0.314) but also the greatest scatter, with points distributed across the entire plot area. The wide vertical spread at any given question-based score indicates high variability in holistic assessment even when structured evaluation is consistent. Notably, GPT-5 shows numerous points in the lower region of the plot (overall assessment < 0.2) even with moderate question-based scores, suggesting that the model's extreme hallucination sensitivity may override positive question-based assessment in determining overall scores.

The mean positions (indicated by stars) reveal the central tendencies: GPT-4o-mini clusters at (0.69, 0.64), GPT-4o at (0.67, 0.59), and GPT-5 at (0.59, 0.18). These positions confirm that while GPT-4o-mini and GPT-4o produce similar question-based assessment, GPT-4o-mini maintains higher and more consistent overall assessment. GPT-5's mean position in the lower-left quadrant reflects both lower question-based assessment and dramatically reduced overall assessment scores.

\begin{figure}[h!]
    \centering
    \includegraphics[width=0.75\textwidth]{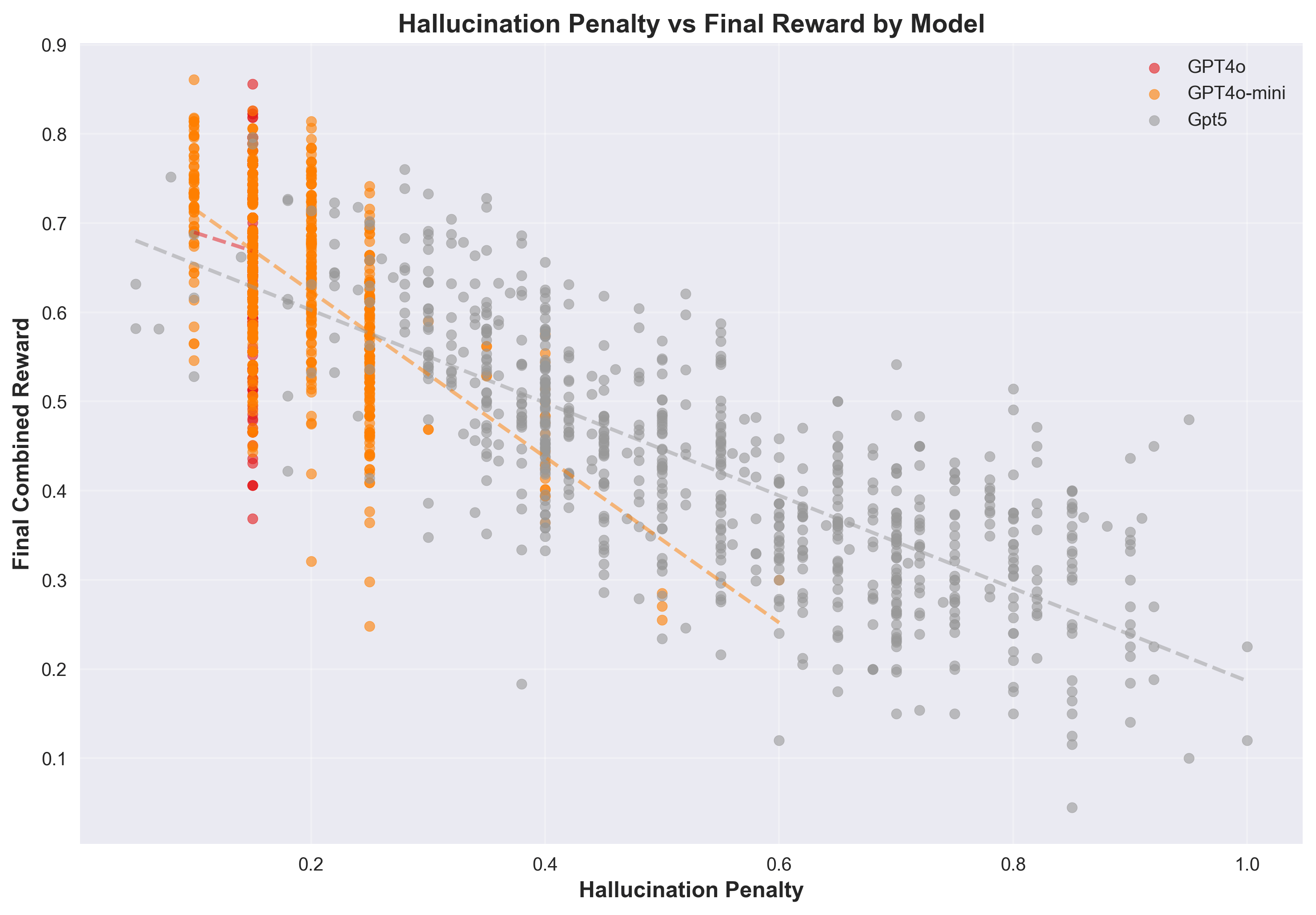}
    \caption{Hallucination Assessment Patterns Reveal Model Validation Strategies. The scatter plot illustrates the relationship between hallucination penalty (x-axis) and final combined reward (y-axis). Trend lines indicate the general relationship between hallucination detection and overall assessment quality. Point clustering patterns reveal systematic differences in validation approaches.}
    \label{fig:hallucination}
\end{figure}

\paragraph{Hallucination Assessment Analysis (Figure~\ref{fig:hallucination})}

Figure~\ref{fig:hallucination} provides critical insights into how hallucination detection influences overall assessment scores across models. GPT-4o (red points) and GPT-4o-mini (orange points) demonstrate remarkably similar patterns, with points densely clustered in the low-hallucination (x < 0.3), high-reward (y > 0.5) region. Both models show strong negative correlations between hallucination penalty and final reward, with concentrated vertical bands at specific hallucination levels (particularly prominent at x $\approx$ 0.15). This clustering suggests systematic hallucination assessment with discrete penalty levels rather than continuous variation.

GPT-4o-mini's orange cluster shows the tightest concentration, with most points falling within x $\in$ [0.1, 0.2] and y $\in$ [0.5, 0.85]. The vertical banding pattern indicates that the model applies standardized hallucination penalties, with three distinct clusters visible at approximately x = 0.10, 0.15, and 0.20. This discretization of hallucination assessment aligns with GPT-4o-mini's overall systematic approach to evaluation.

GPT-5 (gray points) presents a dramatically different pattern, with points distributed across the entire hallucination spectrum (x $\in$ [0, 1.0]) and showing a strong negative correlation between hallucination penalty and reward. The trend line's steep negative slope indicates that hallucination detection dominantly influences GPT-5's final assessment scores. The wide horizontal spread of gray points suggests that GPT-5 assigns highly variable hallucination penalties to similar content, potentially reflecting oversensitivity to perceived inaccuracies.

The density patterns reveal important characteristics: GPT-4o and GPT-4o-mini show high point density in their optimal regions, indicating consistent behavior across samples. GPT-5's points form a diffuse cloud with multiple density centers, suggesting multimodal evaluation behavior where the model alternates between different hallucination detection strategies. The presence of GPT-5 points at extreme hallucination penalties (x > 0.9) with corresponding low rewards (y < 0.2) indicates cases where the model essentially rejects descriptions as fundamentally flawed.

The intersection of trend lines occurs at approximately (0.5, 0.45), suggesting a critical threshold where hallucination penalties begin to dominate assessment scores. Models to the left of this intersection (GPT-4o and GPT-4o-mini) maintain positive assessment despite hallucination detection, while GPT-5's distribution centered around this intersection indicates a fundamental tension between recognition of accurate content and hypersensitivity to potential errors.

\subsection{Case Study: Object Misidentification and Model Response Patterns}

\begin{figure}[h!]
    \centering
    \includegraphics[width=0.8\textwidth]{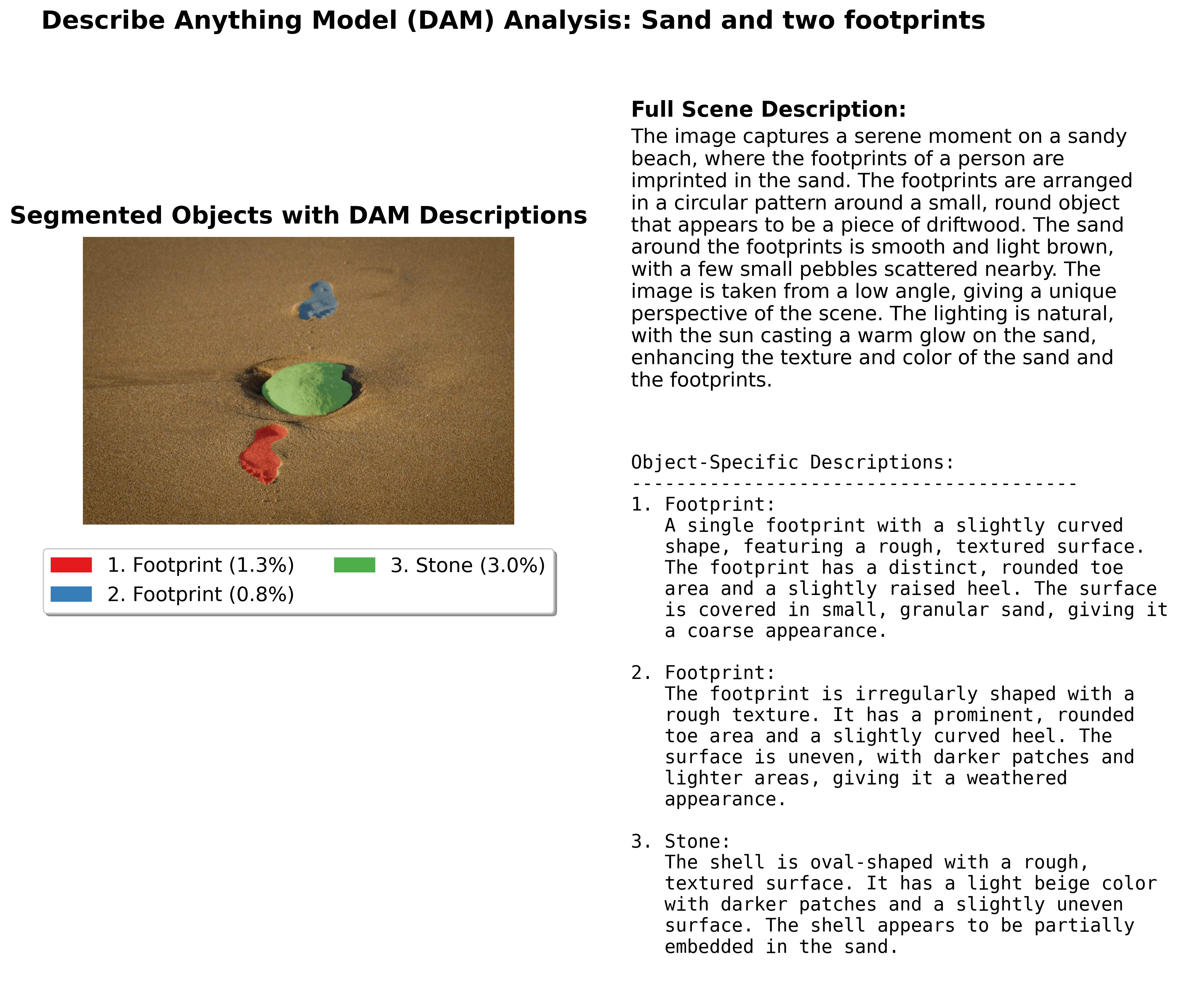}
    \caption{Sample 4 from the DataSeeds.AI dataset showing two footprints in sand with a stone between them, photographed from a high angle in a desert setting. The NVIDIA DAM model critically misidentified the stone as "driftwood" or "shell" and relocated the desert scene to a beach, providing a natural experiment for analyzing differential error weighting across GPT evaluation models. Assessment scores ranged from GPT-4o-mini's lenient 0.41 to GPT-5's severe 0.045, illustrating the extreme variance in evaluation approaches when confronting clear description errors. The percentage beside each object label indicates the coverage percentage of the object with respect to the entire image.}
    \label{fig:case_study_sample4}
\end{figure}

A particularly illuminating example from our dataset demonstrates both the challenges inherent in vision-language description generation and the diverse evaluation strategies employed by different GPT models when confronted with clear contradictions. Figure \ref{fig:case_study_sample4}, titled "Sand and two footprints," provides a natural experiment for understanding how evaluation models respond to fundamental misidentification errors.

\paragraph{Ground Truth and DAM Output}
The human annotations for this image were concise and specific:
\begin{itemize}
    \item \textit{Image description:} "Apparently the picture in desert. Sand, two footprints and a stone between the two steps."
    \item \textit{Scene description:} "High angle scene shows sand with two footprints, a sunny day, between the two footprints there is a buried stone in the sand."
\end{itemize}

The NVIDIA DAM model, however, produced a different interpretation, making several critical errors that cascaded through the entire description. The model incorrectly identified the stone as either "a piece of driftwood" or "a shell," relocated the scene from a desert to "a sandy beach," described the footprints as "arranged in a circular pattern" rather than simply two footprints with an object between them, and specified a "low angle" perspective contradicting the human-annotated high angle view. Additionally, the model introduced unsupported elements such as "small pebbles scattered nearby" and provided excessive detail about textures and colors not present in the human references.

\begin{table}[h!]
\centering
\caption{Quantitative Assessment Results for Sample 4: Object Misidentification Case}
\begin{tabular}{lccc}
\toprule
\textbf{Assessment Metric} & \textbf{GPT-4o} & \textbf{GPT-4o-mini} & \textbf{GPT-5} \\
\midrule
\multicolumn{4}{l}{\textit{Holistic Assessment Components}} \\
Accuracy Score & 0.75 & 0.85 & 0.30 \\
Completeness Score & 0.80 & 0.70 & 0.40 \\
Hallucination Penalty & 0.25 & 0.15 & 0.85 \\
Overall Assessment & 0.52 & 0.64 & 0.00 \\
\midrule
\multicolumn{4}{l}{\textit{Question-Based Assessment}} \\
Positive Questions Generated & 5 & 10 & 10 \\
Negative Questions Generated & 5 & 10 & 10 \\
Positive Score (raw) & 0.5 & 2.0 & 1.5 \\
Negative Score (raw) & 1.0 & 3.0 & 0.0 \\
Positive Score (normalized) & 0.10 & 0.20 & 0.15 \\
Negative Score (normalized) & 0.20 & 0.30 & 0.00 \\
Question-Based Reward & 0.15 & 0.25 & 0.075 \\
\midrule
\textbf{Final Combined Reward} & \textbf{0.31} & \textbf{0.41} & \textbf{0.045} \\
\bottomrule
\end{tabular}
\label{tab:sample4_scores}
\end{table}

\paragraph{Differential Error Emphasis Across Models}

Table~\ref{tab:sample4_scores} reveals striking differences in how each model weighted various errors. GPT-4o-mini, despite recognizing the errors, assigned the highest scores across most dimensions (accuracy: 0.85, overall assessment: 0.64), demonstrating its characteristic leniency. GPT-4o showed more balanced assessment with moderate scores (accuracy: 0.75, overall assessment: 0.52), while GPT-5 applied extreme penalties, particularly for hallucinations (0.85), resulting in a zero overall assessment score.

\begin{table}[h!]
\centering
\caption{Key Error Types Identified and Emphasized by Each GPT Model}
\begin{tabular}{p{3.5cm}ccc}
\toprule
\textbf{Error Type} & \textbf{GPT-4o} & \textbf{GPT-4o-mini} & \textbf{GPT-5} \\
\midrule
\multicolumn{4}{l}{\textit{Object Misidentification}} \\
Stone → Shell/Driftwood & Moderate & Noted & Critical \\
\midrule
\multicolumn{4}{l}{\textit{Setting Errors}} \\
Desert → Beach & Moderate & Minor & Critical \\
High angle → Low angle & Noted & Minor & Critical \\
\midrule
\multicolumn{4}{l}{\textit{Spatial Arrangement}} \\
Two footprints → Circular pattern & Noted & Minor & Critical \\
Between → Around & Moderate & Noted & Critical \\
\midrule
\multicolumn{4}{l}{\textit{Added Elements}} \\
Pebbles (not mentioned) & Minor & Overlooked & Critical \\
Excessive texture details & Overlooked & Overlooked & Moderate \\
\midrule
\multicolumn{4}{l}{\textit{Correctly Identified Elements}} \\
Sunny day/lighting & Confirmed & Confirmed & Confirmed \\
Sand presence & Confirmed & Confirmed & Confirmed \\
Footprints existence & Confirmed & Confirmed & Confirmed \\
\bottomrule
\end{tabular}
\begin{tablenotes}
\small
\item Critical = Major penalty applied; Moderate = Some penalty; Noted = Acknowledged but minimal penalty; Minor = Little to no penalty; Overlooked = Not addressed
\end{tablenotes}
\label{tab:error_emphasis}
\end{table}

The models' explanatory text reveals their distinct evaluation philosophies when confronting these errors:

\textbf{GPT-4o-mini} acknowledged the errors but maintained a generous interpretation: "The AI description is largely accurate, capturing the essence of the scene with details about the footprints, sand texture, and lighting. However, it introduces a circular arrangement of footprints and a specific piece of driftwood, which are not mentioned in the human references, leading to a slight inaccuracy in spatial relationships." This explanation demonstrates GPT-4o-mini's tendency to emphasize what the DAM got right rather than penalizing errors severely.

\textbf{GPT-4o} provided a more balanced critique: "The AI description provides a detailed and vivid depiction of the scene, capturing the serene beach setting and the arrangement of footprints. However, there are discrepancies in accuracy, such as the mention of a piece of driftwood, which is not supported by the human references that describe a stone." This model acknowledged both strengths and weaknesses but still maintained moderate scoring despite recognizing fundamental errors.

\textbf{GPT-5} delivered the harshest assessment: "The AI description captures the general idea of sand, footprints, and sunny lighting, and it provides textural detail. However, it conflicts with key human-reported facts: it sets the scene on a beach (not desert), claims a low angle (humans say high angle), describes multiple footprints arranged in a circular pattern (humans report two footprints), and misidentifies the central object as driftwood or a shell rather than a stone between the two footprints." GPT-5's explanation systematically catalogued every error, treating each as a critical failure rather than a minor discrepancy.

\paragraph{Question Generation Strategies}

The question generation patterns further illuminate model-specific evaluation strategies. GPT-4o generated only 5 positive and 5 negative questions, focusing on core elements with efficient coverage. GPT-4o-mini doubled this to 10 questions each, providing more detailed coverage but potentially diluting the impact of critical errors among minor details. GPT-5 also generated 10 questions each but included highly specific phrases such as "berried stone" (likely interpreting "buried stone"), demonstrating its tendency toward exhaustive and literal evaluation criteria that may become overly rigid.

Particularly revealing is the fact that all three models correctly identified and credited the sunny day aspect (scoring +1 on this dimension) while applying vastly different penalties for the beach/desert confusion. This inconsistency within individual model evaluations suggests that current AI evaluators may struggle with hierarchical error importance, treating all correctly identified details as equally valuable regardless of their centrality to the scene's fundamental characteristics.

\paragraph{Implications of the Case Study}

This single example crystallizes our broader findings about evaluation personalities. GPT-4o-mini's final combined reward of 0.41 despite obvious errors exemplifies its systematic leniency and emphasis on partial credit. GPT-4o's moderate score of 0.31 reflects its balanced approach that recognizes errors without excessive penalization. GPT-5's near-zero score of 0.045 demonstrates how its extreme hallucination sensitivity can completely override any positive aspects of a description, potentially making it too harsh for practical evaluation scenarios where some tolerance for interpretation is necessary.

The case also reveals a fundamental challenge in AI evaluation: the difficulty of establishing appropriate error hierarchies. While misidentifying a central object (stone vs. shell) and relocating an entire scene (desert vs. beach) represent fundamental errors, the models showed varying abilities to weight these appropriately against correctly identified peripheral details. This suggests that future evaluation systems may need explicit hierarchical error frameworks that distinguish between critical misidentifications and minor embellishments.

\subsection{Case Study: Using an independent question generator}

To isolate the evaluation behavior of the GPT models from their question-generation capabilities, 
we conducted a case study using an independent, high-capability model Gemini 2.5 Pro \citep{google2025gemini} 
as a standardized question generator. 

In this experiment, Gemini 2.5 Pro synthesized a single, consistent set of positive and negative questions 
for all 762 samples based on the human ground truth descriptions. Subsequently, each GPT model (GPT-4o, 
GPT-4o-mini, and GPT-5) was tasked with evaluating the DAM-generated descriptions using only this externally-provided 
question set. This decouples the evaluation process, allowing us to measure the pure judgment tendencies of each 
GPT model when presented with identical assessment criteria. The results, summarized in Table~\ref{tab:gemini_question_results}, reveal the distinct "evaluator personalities" 
of the GPT models when their performance is normalized against a common set of questions. 

\begin{table}[h!]
\centering
\caption{GPT Model Evaluation Scores Using Questions Generated by Gemini 2.5 Pro. 
The table displays the mean and standard deviation, presented as percentages, for each metric.}
\label{tab:gemini_question_results}
\begin{threeparttable}
\begin{tabular}{lccc}
\toprule
\textbf{Assessment Metric} & \textbf{GPT-4o} & \textbf{GPT-4o-mini} & \textbf{GPT-5} \\
\midrule
Positive Score & 48.08\% ($\pm$23.85\%) & \textbf{59.59\% ($\pm$22.16\%)} & 40.67\% ($\pm$24.77\%) \\
Negative Score & \textbf{90.89\% ($\pm$16.35\%)} & 86.04\% ($\pm$18.06\%) & 88.97\% ($\pm$20.03\%) \\
Overall Reward & 69.48\% ($\pm$16.06\%) & \textbf{72.82\% ($\pm$16.21\%)} & 64.82\% ($\pm$17.24\%) \\
\bottomrule
\end{tabular}
\end{threeparttable}
\end{table}

The findings from this controlled experiment confirm and clarify the evaluation personalities identified earlier and discussed in Sections \ref{model_assessmet_per} and \ref{model_personalities}. 
GPT-4o-mini emerges as the most lenient or optimistic evaluator, consistently assigning the highest scores for both 
Positive Score (59.59\%) and Overall Reward (72.82\%). Conversely, GPT-5 acts as the most stringent evaluator, 
assigning the lowest scores in these categories. GPT-4o positions itself as a balanced, middle-ground assessor, with 
its scores consistently falling between the other two models. All three models demonstrate relatively similar variance 
levels, suggesting that while their standards for judgment differ, each applies its own standard with a comparable 
degree of consistency. 

\begin{figure}[h!]
\centering
\includegraphics[width=0.75\textwidth]{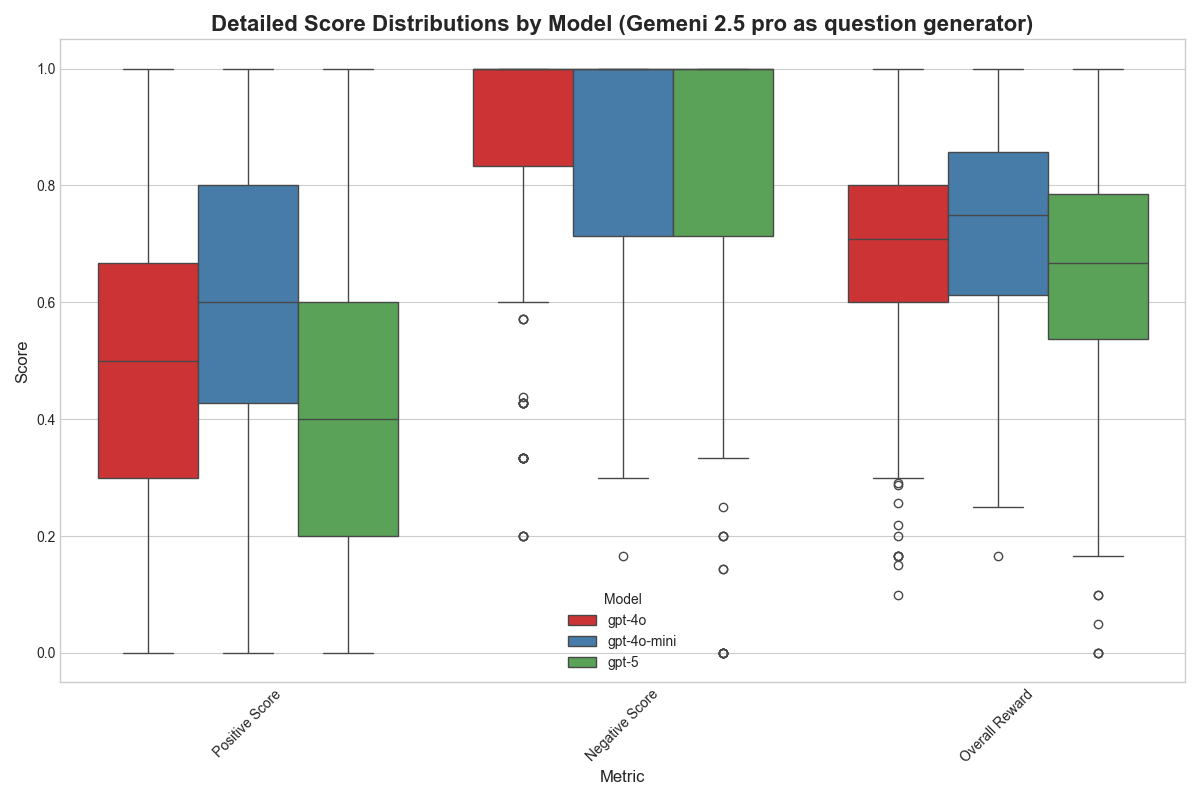}
\caption{Detailed Score Distributions by Model when using Gemini 2.5 Pro as a unified question generator. 
This visualization highlights the different evaluation tendencies of each GPT model when assessing against the same criteria.}
\label{fig:gemini_boxplot}
\end{figure}

The box plot in Figure~\ref{fig:gemini_boxplot} provides a clear visual representation of these distinct evaluation 
patterns. The distribution for `Positive Score` shows the most significant separation between the models, confirming 
that the primary difference in their evaluation strategy lies in how strictly they validate the presence of correct 
information. GPT-5's entire distribution is shifted downwards, indicating a fundamentally harsher standard. 

In contrast, the `Negative Score` distributions are clustered near the top for all models, with median scores at or 
near 1.0. This indicates a strong consensus and high proficiency among all models in identifying when a description 
correctly avoids adding erroneous details. The separation in `Overall Reward` is a direct consequence of the 
disagreement on positive scoring, reinforcing the conclusion that each model possesses a unique and consistent 
"evaluator personality."


\subsection{Question Similarity Results}

To investigate the consistency of evaluation question generation across different LLM architectures, 
we conducted a comprehensive semantic similarity analysis. This analysis compares questions generated 
by the three GPT models against each other and against those generated by an independent, external model---Gemini 2.5 Pro. 
This approach allows us to distinguish between patterns common to the GPT family and those that differ across 
foundational model architectures. 

\begin{figure}[h!]
\centering
\includegraphics[width=\textwidth]{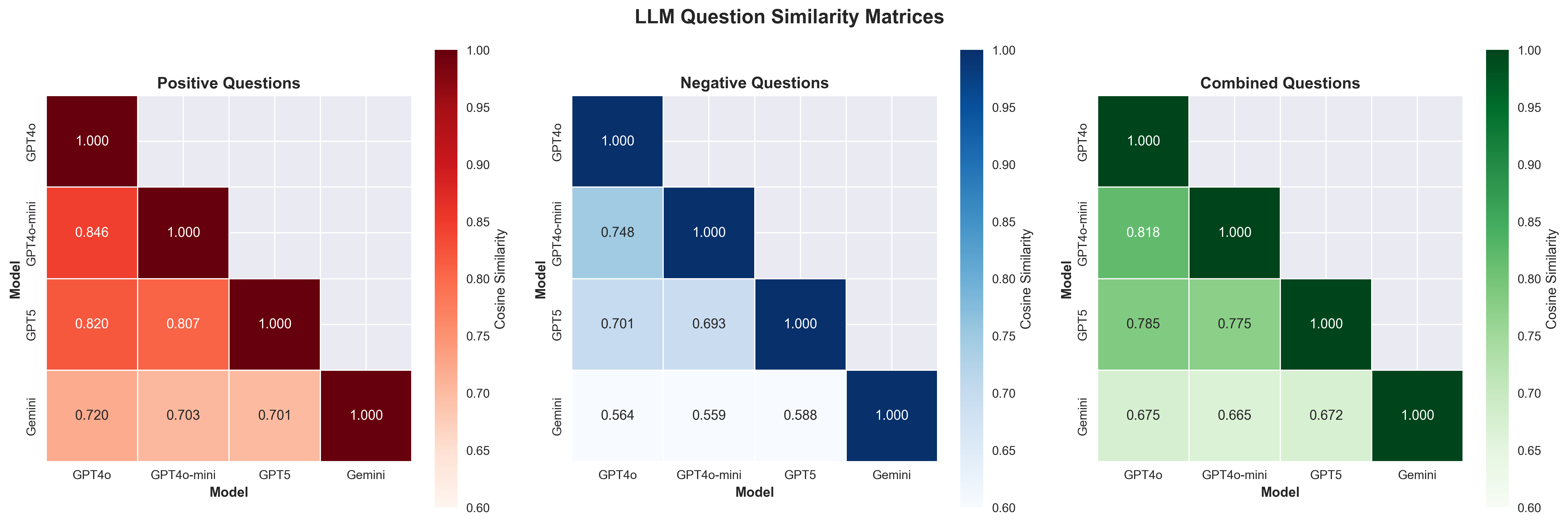}
\caption{Question similarity heatmaps showing semantic similarity between questions generated by different model pairs for positive, negative, and combined cases. We observe a higher degree of similarity among the GPT models compared to Gemini.}
\label{fig:heatmaps_gemini}
\end{figure}

The analysis reveals two primary findings: a clear distinction between the question-generation strategies of the 
GPT family and Gemini, and a universal asymmetry in how all models approach positive versus negative questions. 

\paragraph{Cross-Family Divergence: GPT vs. Gemini:}
Figure~\ref{fig:heatmaps_gemini} illustrates a significant "family gap" in question-generation strategies. The GPT models (GPT-4o, GPT-4o-mini, GPT-5) form a high-similarity cluster, while Gemini stands apart as an 
outlier with consistently lower similarity scores when compared to any GPT model. 

For \textbf{positive questions} (left panel), intra-GPT similarities are very high, ranging from 0.807 to 0.846. 
However, Gemini's similarity to any of the GPT models is markedly lower, hovering around 0.701--0.720. 
This indicates that while the GPT models have converged on a shared understanding of what correct details to verify, 
Gemini employs a distinctly different strategy for positive confirmation. 

This divergence is even more pronounced for \textbf{negative questions} (center panel). The intra-GPT similarities (0.693--0.748) are notably higher than Gemini's similarity to the GPT models (0.559--0.588). A similarity score in the 0.5 range suggests only a moderate conceptual overlap. This implies that Gemini's approach to identifying potential errors, hallucinations, and incorrect details is fundamentally different from the methods 
developed within the GPT family. 

\begin{figure}[h!]
\centering
\includegraphics[width=0.75\textwidth]{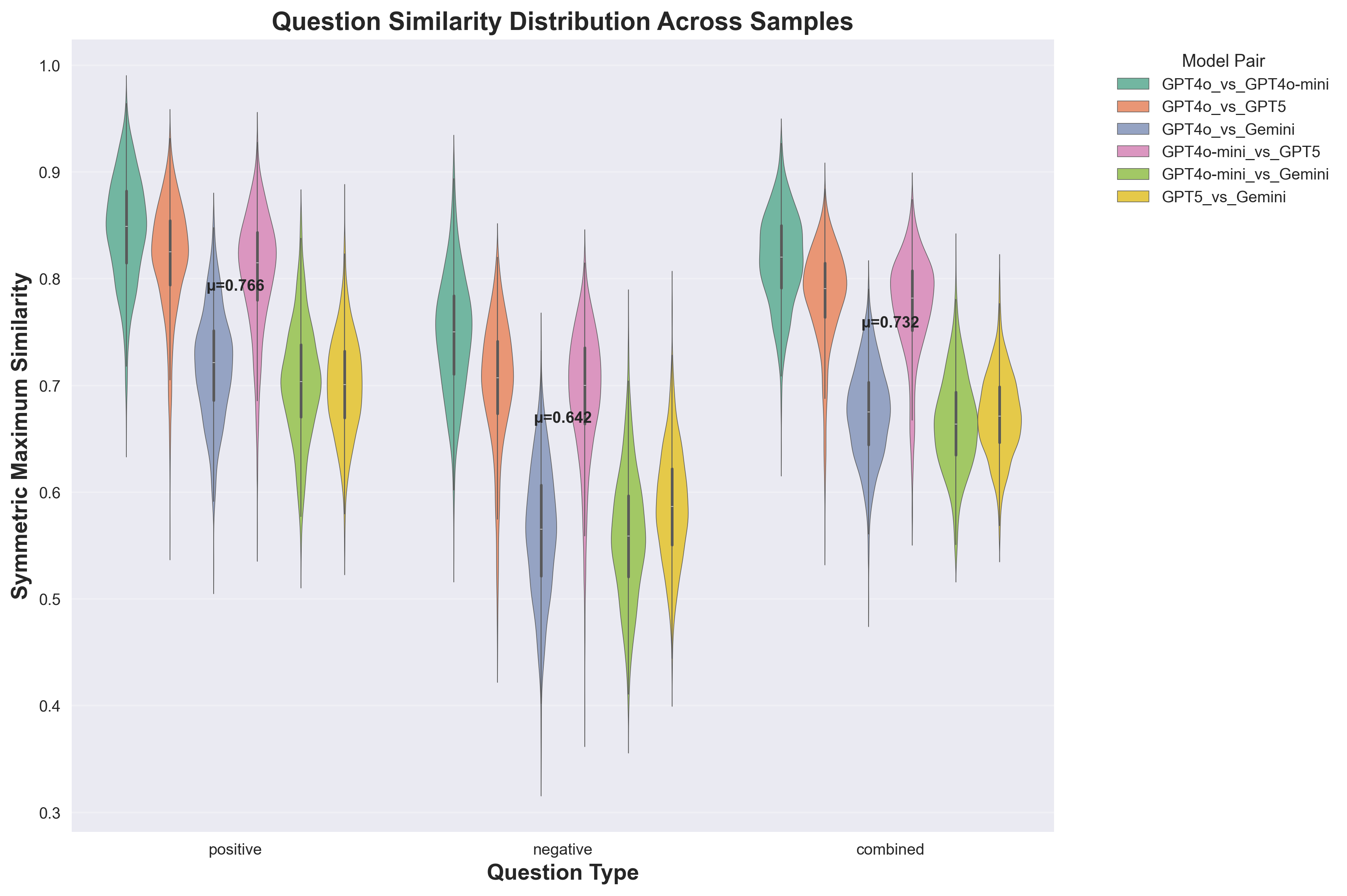}
\caption{Question Similarity Distribution showing violin plots of similarity scores between model pairs for positive, negative, and combined questions. The plots for Gemini pairs are visibly lower and wider, indicating less similarity and more variance.}
\label{fig:violin_plot_gemini}
\end{figure}

\paragraph{Universal Asymmetry: Positive vs. Negative Questions}

Despite the cross-family differences, a universal pattern holds true for all four models: 
they are significantly more aligned on positive verification questions than on negative exclusion questions. 

\begin{figure}[h!]
\centering
\includegraphics[width=0.9\textwidth]{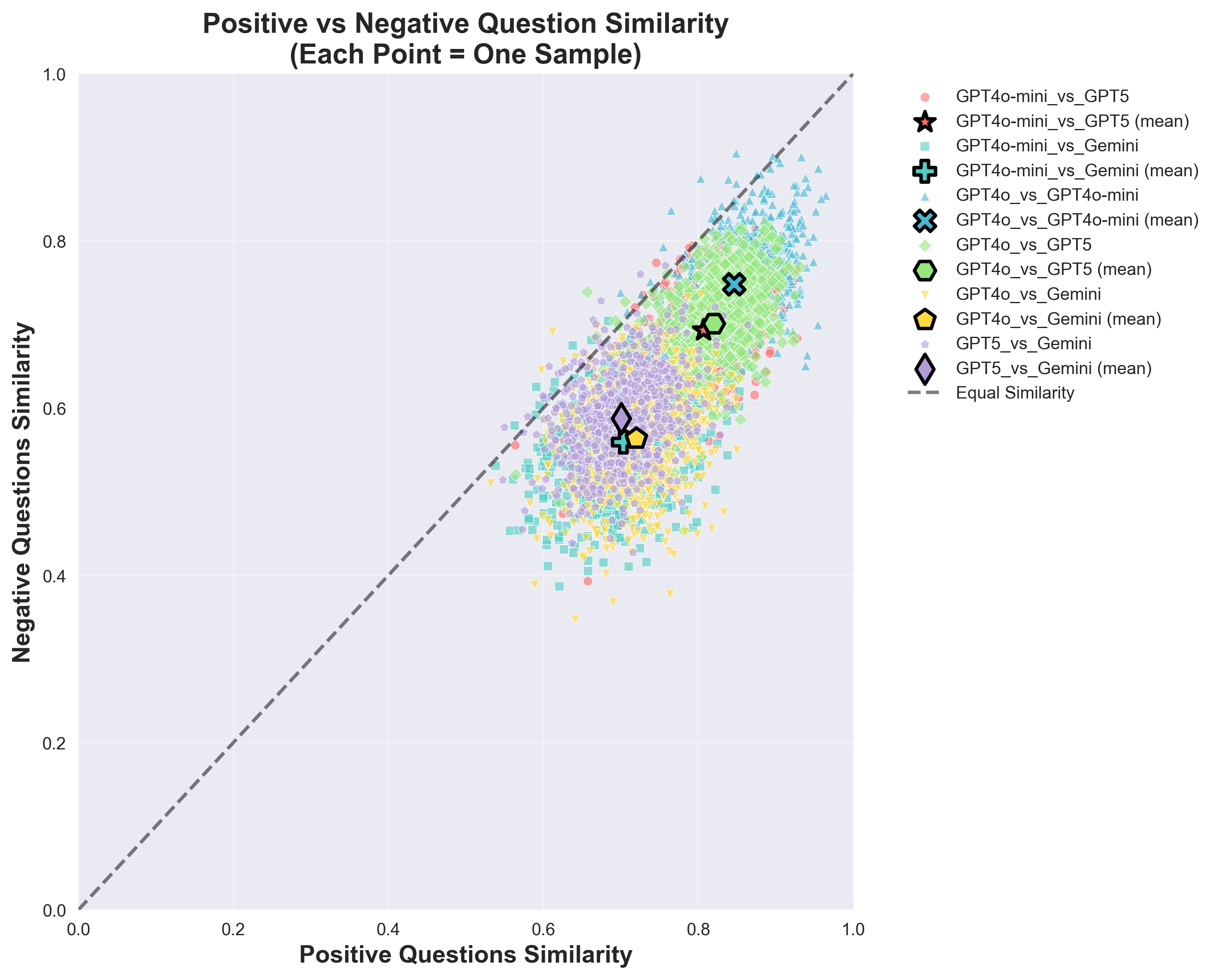}
\caption{Positive vs. Negative Question Similarity, including Gemini pairs. The systematic positioning of all points below the diagonal line confirms that positive question similarity consistently exceeds negative question similarity across all model pairs and samples.}
\label{fig:pos_neg_similarity_gemini}
\end{figure}

Figure~\ref{fig:violin_plot_gemini} visualizes these distributional differences. The distributions for positive 
questions are narrower and concentrated at higher similarity values, indicating a broad consensus on how to confirm 
correct details. In contrast, the distributions for negative questions are wider and centered at lower values. 
This suggests that identifying what is \textit{correct} is a more convergent task, while identifying what might 
be \textit{incorrect} is a more divergent, creative, and model-dependent process. Figure~\ref{fig:pos_neg_similarity_gemini} reinforces this finding at a sample-by-sample level. 
The dense clustering of points below the diagonal line of equality is a powerful illustration that this 
positive-negative asymmetry is not an artifact of averaging but a consistent phenomenon present in nearly 
every single evaluation. 

\paragraph{Implications for Evaluation Reliability}

These findings have significant implications for designing robust, multi-model evaluation frameworks. 
The high intra-family similarity within GPT models suggests they can be used interchangeably for positive 
verification with predictable results. However, their lower agreement on negative questions---and Gemini's 
completely distinct approach---highlights a critical challenge. 

Error and hallucination detection are highly dependent on the model's architecture and training. 
Relying on a single model for this task may overlook entire categories of errors that another model 
might prioritize. Therefore, for comprehensive and unbiased error detection, employing a diverse ensemble 
of evaluators, including models from different architectural families like Gemini, is crucial. 

\section{Discussion}

Our analysis of 762 vision-language descriptions reveals fundamental insights into AI evaluation behavior that challenge assumptions about the relationship between general capability and assessment competence.

\subsection{The Evaluation Paradox}

The emergence of three distinct evaluation personalities demonstrates that assessment behavior represents a complex interplay between architecture, training objectives, and emergent strategies. GPT-4o-mini's near-zero variance ($\pm$0.43\%) in overall assessment suggests convergence on a fixed evaluation function operating independently of content variation. This extraordinary consistency implies the model has developed an abstract representation of quality transcending specific domains.

Conversely, GPT-5's evaluation behavior despite superior general intelligence (94.6\% AIME 2025, 84.2\% MMMU) reveals that optimizations for broad capability introduce instabilities when models perform assessment tasks requiring consistent criteria application. The model's extreme hallucination vigilance (54.10\% penalty rate) coupled with high variance suggests sophisticated contextual reasoning that, while potentially more nuanced, reduces systematic reliability in its role as an evaluator.

GPT-4o exhibits pronounced asymmetry in its evaluation patterns, achieving 92.21\% accuracy when detecting errors but only 41.98\% when confirming correct information. This behavior pattern not a measure of GPT-4o's quality but rather its approach to assessment may reflect how error correction prevalence in training data shapes evaluation strategies, creating implicit biases toward negative assessment when models act as judges. 

The controlled experiment using Gemini 2.5 Pro as an independent question generator further validates these findings. When presented with identical evaluation criteria, the three GPT models maintain their distinct personalities: GPT-4o-mini remains the most lenient (72.82\% overall reward), GPT-5 the most stringent (64.82\%), and GPT-4o balanced between them (69.48\%). This persistence of evaluation patterns when question generation is controlled confirms that these "personalities" are inherent model properties rather than artifacts of how models formulate assessment criteria. The remarkable consistency of these patterns across experimental conditions strengthens our conclusion that evaluation behavior represents a fundamental characteristic of model architecture and training philosophy.

\subsection{Universal Biases and Architectural Trade-offs}

The consistent 2:1 bias favoring negative assessment across all GPT models reveals fundamental characteristics of this model family's training paradigms. However, our cross-family analysis demonstrates this pattern is not universal across AI architectures. The semantic similarity analysis reveals that GPT models achieve high agreement when generating positive verification questions (80.7--84.6\% similarity) but lower agreement on negative detection questions (69.3--74.8\%), a pattern that persists but differs when compared with Gemini's question generation approach.

Our findings reveal fundamental trade-offs in evaluation design. GPT-4o-mini's independence between assessment methods (r = -0.014) suggests architectural separation that prevents error contamination across evaluation dimensions. GPT-5's moderate correlation between methods indicates more integrated evaluation capturing complex interdependencies, but with reduced reliability. The question similarity analysis reveals striking family-specific patterns: while GPT models form a high-similarity cluster (intra-family similarities of 0.78--0.85 for combined questions), Gemini exhibits markedly different question generation strategies (0.67--0.68 similarity to GPT models). This cross-family divergence is particularly pronounced for negative questions, where Gemini's similarity to GPT models drops to 0.56--0.59, suggesting fundamentally different approaches to error detection.

The discovery that positive question similarity consistently exceeds negative question similarity across all model pairs—including cross-family comparisons—indicates a universal asymmetry in how AI systems conceptualize quality assessment. Models converge on what constitutes correct information but diverge significantly in how they identify potential errors. This suggests that while there may be emerging consensus on quality indicators, error detection remains a more subjective, architecture-dependent process shaped by specific training philosophies and safety objectives.

\subsection{Implications for AI Safety and Future Research}
These systematic biases have critical implications for AI safety, particularly as AI systems increasingly evaluate other AI outputs. The recursive nature of AI development, where models evaluate and train other models, makes understanding and correcting evaluation biases crucial for preventing bias amplification.

GPT-5's extreme hallucination vigilance demonstrates how safety-oriented training can create overcorrection that impairs balanced judgment. This suggests current safety approaches may have unintended consequences for evaluation utility. The high variance in GPT-5's evaluation behavior raises concerns about reliability in safety-critical applications where consistent decision-making is essential.

Future research should prioritize several directions:

\textbf{(1) Expanded cross-family evaluation analysis:} Our demonstration of significant GPT-Gemini divergence establishes that evaluation personalities and biases are family-specific rather than universal. The 2:1 negative-positive bias we observed appears to be a GPT family characteristic, as evidenced by the different patterns exhibited by Gemini. Extending this analysis to additional model families like Claude (Anthropic) and open-source alternatives like Llama would reveal whether each architectural lineage develops distinct evaluation philosophies and whether certain approaches yield more balanced assessment.

\textbf{(2) Architecture-specific evaluation models:} Rather than adapting general-purpose models, developing architectures specifically optimized for evaluation tasks could eliminate the trade-off between general capability and assessment consistency. Our finding that GPT-5's superior general intelligence correlates with inferior evaluation consistency suggests dedicated evaluation architectures may be necessary.

\textbf{(3) Controlled bias investigation:} Our independent question generator methodology provides a template for isolating different aspects of evaluation behavior. Systematic experiments varying training data composition, safety constraints, and reward structures while using standardized evaluation criteria could isolate the sources of family-specific biases.

\textbf{(4) Multi-family ensemble methods:} Given the pronounced cross-family divergence we observed, evaluation frameworks should leverage models from different architectural families rather than different versions within a family. Combining GPT's systematic negative detection with Gemini's distinct error conceptualization and potentially other families' strengths could yield more comprehensive and balanced evaluation systems.

\textbf{(5) Human-AI evaluation alignment:} Understanding whether humans exhibit similar evaluation asymmetries and family-like clustering in their assessment approaches would contextualize whether AI evaluation patterns reflect fundamental assessment challenges or training artifacts. The high agreement on positive verification but divergence on error detection we observed may mirror human evaluation patterns.

The significant divergence between GPT and Gemini evaluation strategies, revealed through both question generation patterns and scoring behaviors, demonstrates that robust AI evaluation requires diverse architectural perspectives rather than variations within a single model family.

\section{Conclusion}

This comprehensive study of 762 vision-language descriptions evaluated by three GPT architectures and validated through cross-family comparison with Gemini reveals that evaluation behavior in AI systems represents a complex phenomenon distinct from general intelligence. Through systematic analysis employing dual-assessment methodologies and controlled experiments with independent question generation, we uncovered three distinct evaluation personalities within the GPT family and demonstrated significant cross-family divergence in assessment strategies.

Our primary contribution demonstrates that evaluation competence does not simply scale with model capability. GPT-4o-mini's unprecedented consistency ($\pm$0.43\% variance), GPT-4o's error detection specialization (92.21\% accuracy), and GPT-5's variable high-threshold assessment (54.10\% hallucination penalty) reveal that architectural advances optimized for general intelligence may compromise systematic assessment capabilities. These personalities persist even when question generation is controlled through an independent model, confirming they are inherent architectural properties rather than methodological artifacts.

The discovery of a 2:1 bias toward error detection over positive confirmation within the GPT family, contrasted with Gemini's distinct evaluation patterns, reveals that assessment biases are family-specific rather than universal across AI architectures. The semantic similarity analysis further demonstrates this divergence: GPT models form a high-similarity cluster (0.78--0.85 intra-family similarity) while Gemini exhibits markedly different strategies (0.56--0.72 cross-family similarity), particularly in error detection approaches. This asymmetry has immediate practical implications: evaluation frameworks must incorporate diverse architectural families rather than relying on variations within a single family to achieve balanced assessment.

For practitioners, our results provide evidence-based guidance for model selection and ensemble design. GPT-4o-mini suits applications requiring systematic, reproducible assessment; GPT-4o excels in quality control contexts; and GPT-5's extreme hallucination vigilance, despite overall limitations, may benefit applications where false information detection is paramount. The controlled experiment using Gemini as question generator validates these recommendations by demonstrating that evaluation personalities persist across experimental conditions. The paradox of GPT-5—achieving state-of-the-art performance on complex reasoning while demonstrating poor evaluation consistency—suggests current optimization approaches inadvertently sacrifice specialized capabilities for general intelligence.

Our methodological contributions, including the dual-assessment framework, semantic similarity analysis, and independent question generator protocol, provide tools for future evaluation research. The finding that models achieve high agreement on positive criteria but diverge on error detection reveals fundamental asymmetries in how AI systems conceptualize quality assessment. The pronounced cross-family divergence between GPT and Gemini approaches demonstrates that different architectural lineages develop distinct evaluation philosophies shaped by their training paradigms.

As AI systems increasingly evaluate other AI outputs, understanding and correcting these family-specific biases becomes crucial for preventing recursive bias amplification. The future of AI evaluation lies not in scaling current approaches but in developing diverse, multi-family evaluation frameworks that leverage complementary strengths across architectural lineages. This study establishes that improving AI evaluation requires dedicated attention to assessment as a distinct capability, separate from but complementary to advances in general intelligence, and necessitates systematic incorporation of diverse architectural perspectives to achieve truly balanced and robust evaluation systems.

\newpage
\section*{Appendices}
\appendix
\section{Prompt templates}\label{APP_prompt_templates}
We incorporate the following prompt when prompting the DAM model to generate descriptions from a given image. The complete FULL\_IMAGE\_PROMPT is designed to align with the outputs that human experts are instructed to produce.

\begin{quote}
\small\ttfamily
FULL\_IMAGE\_PROMPT: '<image> Describe this scene
(at least 20 to 30 words and not more than 80 words):
focus on the overall context, environment, lighting, camera angle
(eye level/high/low/bird's eye), color palette,
photography style, and any text visible in the image.'\\
\\
OBJECT\_PROMPTS: '<image> Describe the masked region in detail.'
\end{quote}

We also used the following prompt for generatingm positive and negative questions form the human experts generated descriptions using GPT models. For a fair comparison all the models are prompted using an identical prompt:

\begin{quote}
\small\ttfamily
Generate highly specific evaluation questions based on these human descriptions:\\
\hspace*{2em}Human Image Description: "\{human\_image\_desc\}"\\
\hspace*{2em}Human Scene Description: "\{human\_scene\_desc\}"\\
\hspace*{2em}Segmented Objects Present: "\{segmented\_objects\}"\\
\\
Create very specific questions that check for exact details mentioned in\\
the human descriptions.\\
Each question should test ONE specific factual detail that can be\\
objectively verified.\\
\\
POSITIVE QUESTIONS - Check if AI mentions these SPECIFIC details\\
that SHOULD be present:\\
1. Extract exact numbers (how many objects? specific quantities?)\\
2. Extract specific object types/names (exact species, object categories)\\
3. Extract specific colors mentioned (exact color names, combinations)\\
4. Extract specific poses/actions/states (flying, sitting, moving, etc.)\\
5. Extract specific spatial relationships (above, below, center, background)\\
6. Extract specific technical details (camera angle, lighting conditions,\\
\hspace*{3em}perspective)\\
7. Extract specific mood/atmosphere descriptors (peaceful, dramatic, calm)\\
8. Extract specific visual elements (background type, composition, patterns)\\
9. Extract specific features mentioned (logos, text, decorative elements)\\
10. Extract specific environmental context (settings, time of day, weather)\\
\\
NEGATIVE QUESTIONS - Check if AI incorrectly adds these\\
SPECIFIC wrong details:\\
1. Wrong quantities (different numbers than stated: 1 vs 2, few vs many)\\
2. Wrong object identification (common misidentifications for these\\
\hspace*{3em}specific objects)\\
3. Wrong colors (opposite or commonly confused colors for these objects)\\
4. Wrong actions/poses (static vs dynamic, different specific activities)\\
5. Wrong spatial arrangements (different positioning than described)\\
6. Wrong technical aspects (different angles, lighting,\\
\hspace*{3em}styles than mentioned)\\
7. Wrong mood/atmosphere (opposite emotional tones: calm vs aggressive)\\
8. Wrong/additional elements not mentioned\\
\hspace*{3em}(extra objects, backgrounds, features)\\
9. Wrong environmental details (different settings, weather, time)\\
10. Impossible or contradictory details (physically impossible combinations)\\
\\
Make questions very specific and testable. Use exact words/phrases\\
from human descriptions.
\end{quote}

\section{DAM inference sample}\label{APP_DAM_inf_sample}
The inference of the DAM model for the example shown in Figure (\ref{fig:sample}) is as follows: 

\begin{quote}
\small\ttfamily
\textbf{Full Image Description:} The image features three small birds perched on a metal railing. The birds are of different colors, with one being black and white, another being brown and white, and the third being brown and black. The birds are positioned in a row, with the black and white bird on the left, the brown and white bird in the middle, and the brown and black bird on the right. The background of the image is blurred, but it appears to be a green field. The image is taken from a low angle, looking up at the birds.\\
\\
\textbf{Object-Level Details:}\\
\textbf{Long-tailed Tits\_0:} The sparrow has a plump body with a mix of soft gray and white feathers. Its head is predominantly gray with a subtle hint of black, and it has a small, pointed beak. The wings are a blend of gray and black, with a distinct black patch near the shoulder. The tail is long and slender, featuring a gradient from gray to black. The legs and feet are dark and sturdy, gripping onto a surface.\\
\\
\textbf{Long-tailed Tits\_1:} The sparrow has a fluffy, light brown body with a slightly darker brown patch on its back. Its wings are black with white tips, and it has a small, pointed beak. The legs are thin and dark, with one leg slightly bent.\\
\\
\textbf{Long-tailed Tits\_2:} The sparrow has a fluffy, white chest and underbelly, with a mix of light brown and black feathers on its back and wings. Its head is predominantly white with a black patch around the eyes and extending to the back of the head. The beak is short and black, and the legs are thin and black.\\
\\
\textbf{Metal structures\_3:} A curved, black metal pole with a smooth surface, featuring a slight bend near the top. The pole has a cylindrical shape and is segmented into two visible parts, with the upper segment slightly thicker than the lower segment. The lower segment has a small, protruding bracket-like structure attached to it.\\
\\
\textbf{Metal structures\_4:} A curved metal bar with a smooth, slightly reflective surface, featuring a 90-degree bend. The bar has a consistent diameter and a dark, weathered appearance with some visible wear and tear.\\
\\
\textbf{Nature Background\_5:} Three small birds with black and white markings on their heads and wings, and a mix of brown and white on their bodies. The birds are perched on a black metal structure, with the middle bird slightly turned to the right. The left bird is facing left, and the right bird is facing right.
\end{quote}

\section{Detailed AI judge Example}\label{APP_Detailed_AI judge_sample}

For completeness, we provide an example of the questions generated by each model for a sample image showing three long-tailed tits standing on metal structures (Figure \ref{fig:sample}). The points beside each question indicates the score assigned to each question given the DAM generated description by the corresponding LLM judge. For the questions generated by Gemini 2.5 pro model we show the points assigned by all the GPT models.

\subsection{Gemeni 2.5 pro Generated Questions}
\textbf{Positive Questions:}
\begin{quote}
\small\ttfamily
- Are there exactly three long-tailed tits present in the image? \textbf{GPT 4o: +1 point, GPT 4o-mini: 1 point, GPT 5: -0.5 point}\\
- Are the birds positioned on what appear to be metal structures? \textbf{GPT 4o: +1 point, GPT 4o-mini: 1 point, GPT 5: 1 point}\\
- Is the background behind the birds blurred, making the subjects stand out? \textbf{GPT 4o: +1 point, GPT 4o-mini: 1 point, GPT 5: 1 point}\\
- Does the overall atmosphere of the image feel 'cute and soft', aided by soft, natural lighting? \textbf{GPT 4o: 0 point, GPT 4o-mini: 0 point, GPT 5: 0 point}\\
- Do the birds' feathers display a muted color palette of pinks, whites, grays, and black? \textbf{GPT 4o: -0.5 point, GPT 4o-mini: 0 point, GPT 5: 0 point}\\
- Is the composition a close-up shot of the three birds? \textbf{GPT 4o: +1 point, GPT 4o-mini: 0 point, GPT 5: 0 point}\\
- Does the blurred background primarily feature shades of green and white? \textbf{GPT 4o: 0 point, GPT 4o-mini: 1 point, GPT 5: 0 point}\\
\end{quote}

\textbf{Negative Questions:}
\begin{quote}
\small\ttfamily
- Are the birds perched on a natural tree branch instead of a metal structure? \textbf{GPT 4o: +1 point, GPT 4o-mini: +1 point, GPT 5: +1 point}\\
- Is the background in sharp, clear focus? \textbf{GPT 4o: +1 point, GPT 4o-mini: +1 point, GPT 5: +1 point}\\
- Is the lighting in the image harsh and direct, creating strong shadows? \textbf{GPT 4o: +1 point, GPT 4o-mini: +1 point, GPT 5: +1 point}\\
- Are there more than three birds, or are they a different species like sparrows? \textbf{GPT 4o: -1 point, GPT 4o-mini: +1 point, GPT 5: -1 point}\\
- Is the image a wide shot, showing a large landscape with the birds as small elements? \textbf{GPT 4o: +1 point, GPT 4o-mini: +1 point, GPT 5: +1 point}\\
- Are the dominant colors in the image bright and primary (e.g., vibrant reds, blues, yellows) rather than muted pinks and grays? \textbf{GPT 4o: +1 point, GPT 4o-mini: +1 point, GPT 5: +1 point}\\
- Are the birds captured in mid-flight rather than standing still? \textbf{GPT 4o: +1 point, GPT 4o-mini: +1 point, GPT 5: +1 point}\\
\end{quote}

\subsection{GPT-4o Generated Questions}
\textbf{Positive Questions:}
\begin{quote}
\small\ttfamily
- Does the description mention exactly 3 long-tailed Tits? \textbf{+1 point}"\\
- Does the description specifically identify the birds as 'long-tailed Tits'? - \textbf{0 point}\\
- Does the description mention the colors 'pinks', 'whites', 'grays', and 'black'? \textbf{-0.5} point\\
- Does the description state that the birds are 'standing' on metal structures? \textbf{+1 point}\\
- Does the description mention a 'blurred background' highlighting the birds? \textbf{+1 point}\\
- Does the description specify 'natural, soft lighting' conditions? \textbf{0 point}\\
- Does the description convey a 'cute and soft' atmosphere? \textbf{0 point}\\
- Does the description mention a 'muted urban color palette' against a 'green and white blurred background'? \textbf{0 point}
\end{quote}

\textbf{Negative Questions:}
\begin{quote}
\small\ttfamily
- Does the description incorrectly state there are fewer than 3 or more than 3 birds? \textbf{+1 point}\\
- Does the description misidentify the birds as sparrows, robins, or other species? \textbf{+1 point}\\
- Does the description incorrectly\\
 mention bright or vivid colors instead of the muted palette? \textbf{+1 point}\\
- Does the description incorrectly state the birds \textbf{+1 point}\\
 are flying or moving instead of standing? \textbf{+1 point}\\
- Does the description incorrectly place the birds \textbf{+1 point}\\
 in a forest or rural setting instead of an urban one? \textbf{+1 point}\\
- Does the description incorrectly mention harsh or artificial lighting instead of natural, soft lighting? \textbf{+1 point}\\
- Does the description convey an aggressive or tense atmosphere instead of a cute and soft one? \textbf{+1 point}\\
- Does the description incorrectly add elements like water bodies, mountains, or additional animals not mentioned? \textbf{+1 point}
\end{quote}

\subsection{GPT-4o-mini Generated Questions}

\textbf{Positive Questions:}
\begin{quote}
\small\ttfamily
- Does the description mention exactly 3 long-tailed Tits? \textbf{+1 point}\\
- Does the description specifically identify the objects as 'metal structures'? \textbf{+1 point}\\
- Does the description mention the colors 'pink', 'white', 'gray', and 'black' in the muted urban color palette? +\textbf{0 point}\\
- Does the description state that the long-tailed Tits are standing? \textbf{+1 point}\\
- Does the description indicate that the blurred background is 'green and white'? \textbf{0 point}\\
- Does the description specify that the lighting is 'natural' and 'soft'? \textbf{0 point}\\
- Does the description convey a 'cute' and 'soft' atmosphere? \textbf{0 point}\\
- Does the description mention a 'nature background'? \textbf{+1 point}\\
- Does the description include any specific features like logos or text? \textbf{0 point}\\
- Does the description imply an urban setting? \textbf{0 point} \\
\end{quote}

\textbf{Negative Questions:}
\begin{quote}
\small\ttfamily
- Does the description incorrectly state there are only 2 long-tailed Tits or more than 3? \textbf{+1 point}\\
- Does the description misidentify the birds as another species such as sparrows or finches? \textbf{+1 point}\\
- Does the description incorrectly mention colors like 'blue' or 'red' instead of the specified colors? \textbf{+1 point}\\
- Does the description incorrectly state that the long-tailed Tits are flying instead of standing? \textbf{+1 point}\\
- Does the description suggest that the background is 'bright' or 'distracting' instead of 'blurred'? \textbf{+1 point}\\
- Does the description mention 'harsh' or 'artificial' lighting instead of 'natural' and 'soft'? \textbf{+1 point}\\
- Does the description convey an 'aggressive' or 'tense' atmosphere instead of 'cute' and 'soft'? \textbf{+1 point}\\
- Does the description include additional elements like other birds or objects not mentioned? \textbf{+1 point}\\
- Does the description incorrectly state the setting as a rural area instead of an urban one? \textbf{+1 point}\\
- Does the description include physically impossible details, such as the birds standing on a moving object? \textbf{+1 point}
\end{quote}

\subsection{GPT-5 Generated Questions}

\textbf{Positive Questions:}
\begin{quote}
\small\ttfamily
- Does the description state there are exactly three long-tailed tits? - \textbf{+1 point}\\
- Does the description specifically identify the birds as "long-tailed tits" (the species name)? - \textbf{+1 point}\\
- Does the description say the birds are "standing"? - \textbf{0 point}\\
- Does the description say the birds are on "metal structures"? - \textbf{+1 point}\\
- Does the description explicitly use the plural phrase "metal structures" (more than one structure)? - \textbf{+1 point}\\
- Does the description place the birds against "a blurred background"? - \textbf{+1 point}\\
- Does the description specify that the background colors are "green and white"? - \textbf{0 point}\\
- Does the description mention "natural, soft lighting"? - \textbf{0 point}\\
- Does the description describe the shot as a "close up picture"? - \textbf{0 point}\\
- Does the description describe the atmosphere as "cute and soft"? - \textbf{0 point}\\
- Does the description mention a "muted urban color palette" of "pinks, whites, grays, and black"? - \textbf{0 point}\\
- Does the description highlight the birds' "colorful feathers"? - \textbf{0 point}\\
- Does the description refer to the birds as "cute little creatures"? - \textbf{0 point}
\end{quote}

\textbf{Negative Questions:}
\begin{quote}
\small\ttfamily
- Does the description incorrectly state there is only 1 bird, 2 birds, or more than 3 birds instead of exactly three? - \textbf{+1 point}\\
- Does the description misidentify the birds as sparrows, chickadees, bushtits, wrens, swallows, or other species instead of long-tailed tits? - \textbf{-1 point}\\
- Does the description claim the birds are flying, in mid-flight, or moving rapidly instead of standing? - \textbf{+1 point}\\
- Does the description say the birds are perched on wood, branches, a tree, or a wooden fence instead of on metal structures? - \textbf{+1 point}\\
- Does the description claim the background is sharp or in focus instead of blurred? - \textbf{+1 point}\\
- Does the description state the background colors are blue, orange, or purple instead of green and white? - \textbf{+1 point}\\
- Does the description mention harsh artificial lighting, flash lighting, or dramatic high-contrast lighting instead of natural, soft lighting? - \textbf{+1 point}\\
- Does the description call it a wide shot, distant shot, or landscape view instead of a close up picture? - \textbf{+1 point}\\
- Does the description describe the mood as dramatic, aggressive, tense, or ominous instead of cute and soft? - \textbf{+1 point}\\
- Does the description add elements like logos, text overlays, signage, or captions that are not mentioned? - \textbf{+1 point}\\
- Does the description place the scene indoors (studio, room, or cage) instead of in natural, soft lighting with a green and white blurred background? - \textbf{+1 point}\\
- Does the description mention rain, snow, fog, nighttime, or sunset lighting conditions that are not stated? - \textbf{+1 point}\\
- Does the description claim there is only one metal structure instead of "metal structures" (plural)? - \textbf{-1 point}\\
- Does the description assert that the background is both blurred and sharply detailed at the same time (a contradiction)? - \textbf{+1 point}
\end{quote}

\subsection{Aggregated judgments}

Table \ref{tab:gpt_detailed_judges_scores} summarizes the aggregated scores assigned by each LLM as judge according to overall assessment and also question based assessment for the sample shown in Figure \ref{fig:sample}. Note that for the Question based scores the corresponding GPT model is used to generate the questions.

\begin{table}[h!]
\centering
\caption{Evaluation Metrics for Different GPT Models as LLM Judges}
\label{tab:gpt_detailed_judges_scores}
\begin{tabular}{lccc}
\toprule
\textbf{Metric} & \textbf{GPT-4o} & \textbf{GPT-4o-mini} & \textbf{GPT-5} \\
\midrule
\multicolumn{4}{l}{\textit{Question-Based Scores}} \\
Positive Total & 2.5 & 4.0 & 5.0 \\
Negative Total & 8.0 & 10.0 & 10.0 \\
Positive Score (Normalized) & 0.313 & 0.400 & 0.385 \\
Negative Score (Normalized) & 1.000 & 1.000 & 0.714 \\
Question-Based Reward & 0.656 & 0.700 & 0.549 \\
\midrule
\multicolumn{4}{l}{\textit{Holistic Assessment}} \\
Accuracy & 0.850 & 0.850 & 0.580 \\
Completeness & 0.700 & 0.700 & 0.550 \\
Hallucination Penalty & 0.150 & 0.150 & 0.470 \\
Overall Assessment Reward & 0.640 & 0.640 & 0.098 \\
\midrule
\textbf{Final Combined Reward} & \textbf{0.650} & \textbf{0.676} & \textbf{0.369} \\
\bottomrule
\end{tabular}
\end{table}

Each LLM judge also provided a detailed explanation of the holistic judgment that it made. Table~\ref{tab:holistic_explanations} shows the detailed analytical descriptions of the judgments.

\begin{table}[h!]
\centering
\caption{Detailed Holistic Judgment Explanations by LLM Judges}
\label{tab:holistic_explanations}
\begin{tabular}{p{2.5cm}p{11.5cm}}
\toprule
\textbf{LLM Judge} & \textbf{Explanation} \\
\midrule
\textbf{GPT-4o} & The AI description is mostly accurate in terms of the number of birds and their general positioning on the railing. It correctly identifies the presence of three birds and provides a detailed account of their colors and positions. However, it inaccurately refers to the birds as 'sparrows' instead of 'long-tailed tits,' which is a factual error. The description also includes detailed object-level descriptions that are not present in the human references, such as specific details about the metal structures and the birds' physical features, which may not be entirely accurate or necessary. In terms of completeness, the AI description captures the main subjects but lacks the mood and atmospheric details present in the human references, such as the 'soft lighting' and 'muted urban color palette.' The hallucination penalty is applied due to the inclusion of unsupported details, such as the specific descriptions of the metal structures and the incorrect identification of the bird species. \\
\midrule
\textbf{GPT-4o-mini} & The AI description is largely accurate, correctly identifying the colors and types of birds, as well as their arrangement on the railing. However, there are minor inaccuracies in the identification of the birds, as they are referred to as sparrows instead of long-tailed tits, which could lead to confusion. The completeness of the description is moderate; it captures the main subjects and some technical aspects like the low angle of the shot but lacks the emotional tone and atmosphere conveyed in the human references. The AI adds some details about the birds' physical characteristics and the metal structures that are not present in the human descriptions, which contributes to a hallucination penalty. Overall, while the AI provides a detailed account, it misses some of the nuanced qualities that human descriptions convey. \\
\midrule
\textbf{GPT-5} & Strengths: The AI correctly notes there are three small birds perched on metal with a blurred green background, broadly aligning with the references. Weaknesses: It repeatedly misidentifies the birds as sparrows rather than long-tailed tits and only lightly reflects the long-tailed trait. It omits key aspects from the references such as the soft, cute mood, natural soft lighting, close-up feel, and the muted palette that includes pinks. The description adds unsupported specifics, including a low angle viewpoint, detailed segmentation and brackets on the metal, precise bird color ordering and wing patterns, and other fine-grained attributes not grounded in the human references. Overall, the result is partially accurate on core scene elements but undermined by species mislabeling and several speculative details, with incomplete coverage of the intended aesthetic and technical cues. \\
\bottomrule
\end{tabular}
\end{table}

\section{Algorithm Details}

\begin{algorithm}
\caption{Complete Evaluation Workflow}
\begin{algorithmic}[1]
\State \textbf{Input:} Dataset D with 762 samples, each containing human descriptions and segmentation masks
\State \textbf{Output:} Assessment scores and patterns for each GPT model
\ForAll{sample $s \in D$}
    \State Generate DAM descriptions for full image and segmented objects
    \ForAll{GPT model $m \in \{GPT-4o, GPT-4o-mini, GPT-5\}$}
        \State Generate positive and negative questions from human descriptions
        \State Perform question-based evaluation on DAM descriptions
        \State Perform holistic assessment (accuracy, completeness, hallucination)
        \State Compute combined scores using weighted formulas
        \State Store results for pattern analysis
    \EndFor
\EndFor
\State Analyze cross-model patterns and biases
\State Generate visualizations and statistical comparisons
\State Return assessment patterns and model evaluation personalities
\end{algorithmic}
\end{algorithm}

\clearpage
\bibliography{files/references}
\bibliographystyle{plainnat}

\end{document}